\documentclass[10pt,twocolumn,letterpaper]{article}
\pdfoutput=1
\usepackage{cvpr}
\usepackage{times}
\usepackage{epsfig}
\usepackage{graphicx}
\usepackage{amsmath}
\usepackage{amssymb}
\usepackage{pdfpages}

\usepackage{algpseudocode} 
\usepackage{amsmath}
\usepackage{enumerate}
\usepackage{verbatim}

\newcommand{\xiao}{\textcolor{black}}

\usepackage[pagebackref=true,breaklinks=true,letterpaper=true,colorlinks,bookmarks=false]{hyperref}
\usepackage[author={DP}, open=true]{pdfcomment}

\cvprfinalcopy 

\def\cvprPaperID{192} 

\ifcvprfinal\fi
\begin{document}

\title{Don't Just Listen, Use Your Imagination: \\ Leveraging Visual Common Sense for Non-Visual Tasks}

\author{Xiao Lin\\
Virginia Tech\\
Blacksburg, VA\\
{\tt\small linxiao@vt.edu}
\and
Devi Parikh\\
Virginia Tech\\
Blacksburg, VA\\
{\tt\small parikh@vt.edu}
}

\maketitle

\begin{abstract}
Artificial agents today can answer factual questions. But they fall short on questions that require common sense reasoning. Perhaps this is because most existing common sense databases rely on text to learn and represent knowledge. 
But much of common sense knowledge is unwritten -- partly because it tends \xiao{not to} be interesting enough to talk about, and partly because some common sense is unnatural to articulate in text. While unwritten, it is not unseen. 
In this paper we leverage semantic common sense knowledge learned from images -- \ie visual common sense -- in two textual tasks: fill-in-the-blank and visual paraphrasing. We propose to ``imagine'' the scene behind the text, and leverage \xiao{visual cues from the ``imagined'' scenes} in addition to textual cues while answering these questions. We imagine the scenes \xiao{as} a visual abstraction. Our approach \xiao{outperforms a} strong text-only baseline on these tasks. Our proposed tasks can serve as benchmarks to quantitatively evaluate progress \xiao{in} solving tasks that go ``beyond recognition''. Our code and datasets will be made publicly available.


\end{abstract}


\vspace{-10pt}
\section{Introduction}
\label{sec:intro}

Today's artificially intelligent agents are good at answering factual questions about our world~\cite{Dang_2007,Ferrucci_2010,Unger_2014}. For instance, Siri\footnote{\url{https://www.apple.com/ios/siri/}}, Cortana\footnote{\url{http://www.windowsphone.com/en-us/how-to/wp8/cortana/meet-cortana}}, Google Now\footnote{\url{http://www.google.com/landing/now/}}, Wolfram Alpha\footnote{\url{http://www.wolframalpha.com/}} \xiao{\etc,} when asked ``How far is the closest McDonald's to me?'', can comprehend the question, mine the appropriate database (\eg maps) and respond with a useful answer. While being good at niche applications or answering factual questions, today's AI systems are far from being sapient intelligent entities. Common sense continues to elude them. 

\begin{figure}
\begin{center}
   \includegraphics[width=1\linewidth]{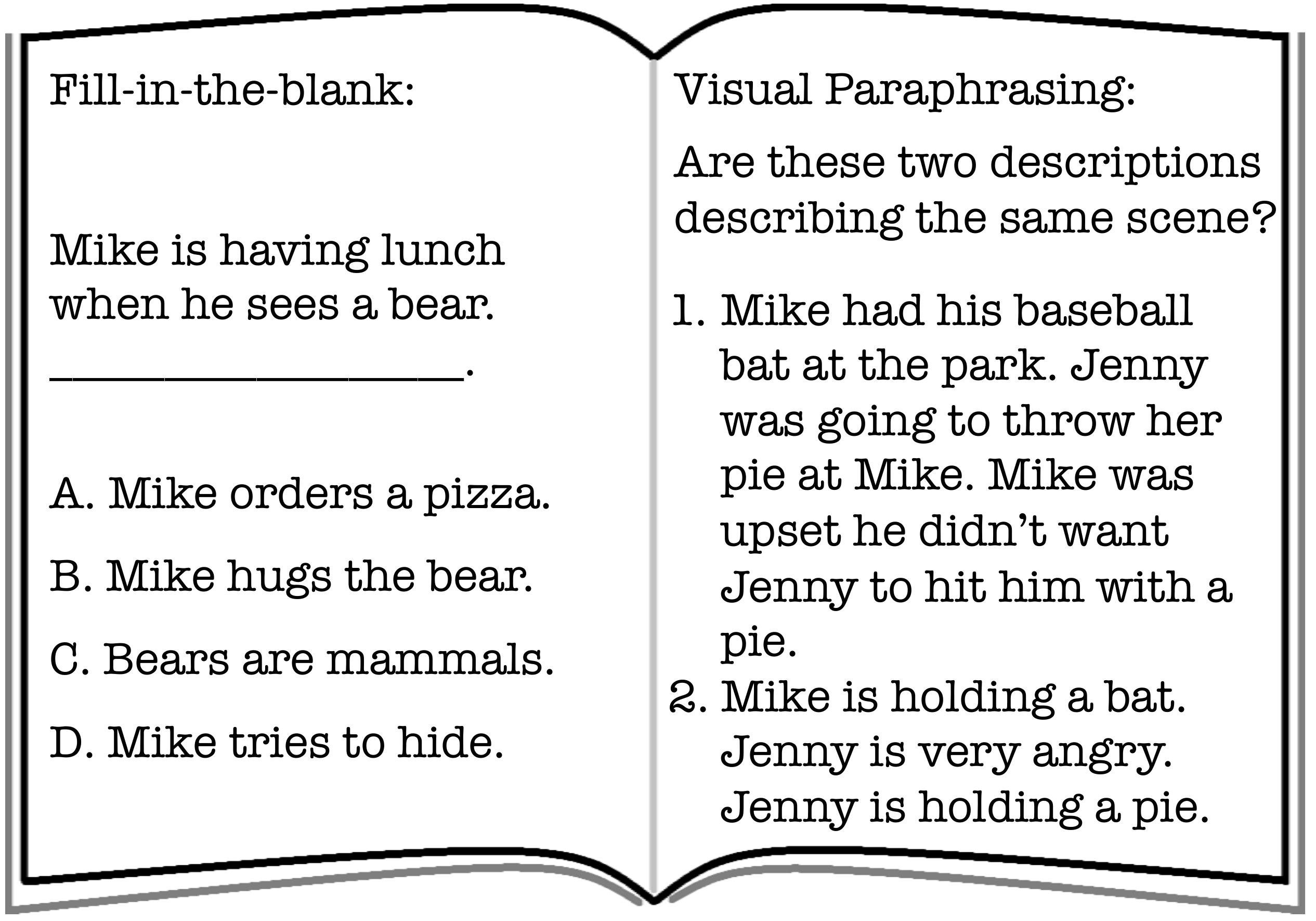}
\end{center}
\vspace{-10pt}
   \caption{We introduce two tasks: fill-in-the-blank (FITB) and visual paraphrasing (VP). While they seem like purely textual tasks, they require some imagination -- visual common sense -- to answer.
   }
   \vspace{-10pt}
\label{fig:teaser}
\end{figure}

Consider a simple fill-in-the-blank task shown in Figure~\ref{fig:teaser} (left). Answering this question requires \xiao{the} common sense that bears are dangerous animals, people like to stay away from and not be noticed by dangerous animals, and hiding is one way of going unnoticed. Similarly, consider the visual paraphrasing question in Figure~\ref{fig:teaser} (right). Answering this question involves common sense that people \xiao{might} throw things when they are angry. Today's systems are unable to answer such questions reliably.

Perhaps this is not surprising. Most existing common sense knowledge bases rely on knowledge described via text -- either mined~\cite{Carlson_2010,Hoffart_2013,DBPedia_2014} or manually entered~\cite{Miller_1995,Singh_2002,Bollacker_2008,Speer_2013}. There are a few short\xiao{-}comings of learning common sense from text. First, it has been shown that people tend \xiao{not to} explicitly talk about common sense knowledge in text~\cite{Gordon_2013}.  Instead, there is a bias to talk about unusual circumstances, because those are worth talking about. Co-occurrence statistics of visual concepts \xiao{mined} from the web has been shown to not generalize to images~\cite{Mensink_2014}. Even when describing images, text is likely to talk about the salient ``foreground'' objects, activities, \etc. But common sense reveals itself even in the ``background''. Second, much of useful common sense knowledge may be hard to describe in text. For instance, the knowledge that ``one person is running after another person'' implies that the first person is facing the second person, the second person is looking in the same direction as the first person, and both people are in running poses, is unnatural (and typically unnecessary) to articulate in text.

Fortunately, much of this common sense knowledge is depicted in our visual world. We call such common sense knowledge that can be learnt from visual data \emph{visual common sense}. By visual common sense we do not mean visual models of commonly occurring interactions between objects~\cite{Divvala_2014} or knowledge of visual relationships between objects, parts and attributes~\cite{Chen_2013,Zhu_2014}. We mean semantic common sense, \eg the knowledge that if one person is running after another person, and the second person turns around, he will see the first person. \xiao{It} can be learnt from visual data but can help \xiao{in} a variety of \xiao{visual \emph{and} non-visual} AI tasks. \xiao{Such} visual common sense is complementary to common sense learnt from non-visual sources.

We argue that the tasks shown in Figure~\ref{fig:teaser} may look like purely text- or language-based tasks on the surface, but they can benefit from visual common sense. In fact, we go further and argue that such tasks can provide exciting new benchmarks to evaluate image understanding ``beyond recognition''. Effectively learning and applying visual common sense to such tasks involves challenges such as grounding language in vision and learning common sense from visual data -- both steps towards deeper image understanding \xiao{beyond} naming objects, attributes, parts, scenes and other image content depicted in the pixels of an image.

In this work we propose two tasks: fill-in-the-blank (FITB) and visual paraphrasing (VP) -- as seen in Figure~\ref{fig:teaser} -- that can benefit from visual common sense. We propose an approach to address these tasks that first ``imagines'' the scene behind the text. It then reasons about the generated scenes using visual common sense, as well as the text using textual common sense, to identify the most likely solution to the task. In order to leverage visual common sense, this imagined scene need not be photo-realistic. It only needs to encode the semantic features of a scene (which objects are present, where, what are their attributes, how are they interacting, \etc.). Hence, we imagine our scenes in an abstract representation of our visual world -- in particular using clipart~\cite{Zitnick_2013_CVPR,Zitnick_2013_ICCV,Fouhey_2014,Antol_2014}.

Specifically, given \xiao{an} FITB task with four options, we generate a scene corresponding to each of the four descriptions that can be formed by pairing the input description with each of the four options. We then apply a learnt model that reasons jointly about text and vision to \xiao{select} the most plausible option. 
Our model essentially uses the generated scene as an intermediate representation to help solve the task.
Similarly, for a VP task, we generate a scene for each of the two descriptions, and apply a learnt joint text and vision model to classify both descriptions as describing the same scene or not. We introduce datasets for both tasks. We show that our imagination-based approach that leverages both visual and textual common sense outperforms the text-only baseline on both tasks. Our datasets and code will be made publicly available.
\section{Related Work}
\label{sec:related_work}

\textbf{Beyond recognition:} Higher-level image understanding tasks go beyond recognizing and localizing objects, scenes, attributes and other image content depicted in the pixels of the image. Example tasks include reasoning about \emph{what} people talk about in images~\cite{Berg_2012}, understanding the flow of time (\emph{when})~\cite{Pickup_2014}, identifying \emph{where} the image is taken~\cite{Hays_2008,Khosla_2014} and judging the intentions of people in images (\emph{why})~\cite{Pirsiavash_2014}. While \xiao{going} beyond recognition, these tasks are fairly niche. Approaches that automatically produce a textual description of images~\cite{Gupta_2008,Farhadi_2010,Kulkarni_2011} or synthesize scenes corresponding to input textual descriptions~\cite{Zitnick_2013_ICCV} can benefit from reasoning about all these different ``W'' questions \xiao{and other high-level information}. They are semantically more comprehensive variations of beyond recognition tasks that test \xiao{high-level} image understanding abilities. However\xiao{,} these tasks are difficult to evaluate~\cite{Kulkarni_2011,Elliott_2014} \xiao{or} often evaluate aspects of the problem that are less relevant to image understanding \eg grammatical correctness of automatically generated descriptions of images. This makes it difficult to use these tasks as benchmarks for evaluating image understanding beyond recognition.

Leveraging visual common sense in our proposed FITB and VP tasks requires qualitatively a similar level of image understanding as in image-to-text and text-to-image tasks. FITB requires reasoning about what else is plausible in a scene given a partial textual description. 
VP tasks on the other hand require us to reason about how multiple descriptions of the same scene could vary. At the same time, 
FITB and VP tasks are multiple-choice questions and hence easy to evaluate. This makes them desirable benchmark tasks for evaluating image understanding beyond recognition.

\textbf{Natural language Q\&A:} Answering factual queries in natural language is a well studied problem in text retrieval. Given questions like ``Through which country does the Yenisei river flow?'', the task is to query useful information sources and give \xiao{a correct} answer for example ``Mongolia'' or ``Russia''. Many systems such as personal assistant applications on phones and IBM Watson \cite{Ferrucci_2010} which won the Jeopardy! challenge have achieved commercial success. There are also established challenges on answering factual questions posed by humans \cite{Dang_2007}, natural language knowledge base queries \cite{Unger_2014} and even university entrance exams \cite{Penas_2014}.  The FITB and VP tasks we study are not about facts, but common sense questions.

\textbf{Leveraging common sense:} Common sense is an important element in solving many beyond recognition tasks, since beyond recognition tasks tend to require information that is outside the boundaries of the image.  It has been shown that learning and using \emph{non-visual} common sense (\ie common sense learnt from non-visual sources) benefits physical reasoning~\cite{Hamrick_2011,Zheng_2013}, reasoning about intentions~\cite{Pirsiavash_2014} and object functionality~\cite{Zhu_2014}. One instantiation of visual common sense that has been leveraged in the vision community in the past is the use of contextual reasoning for improved recognition~\cite{Gupta_2008,Divvala_2009,Grabner_2011,Fouhey_2012,Hedau_2012,Zhu_2014}. 
In this work, we explore the use of visual common sense for seemingly non-visual tasks through ``imagination'', \ie generating \xiao{scenes}. 

\textbf{Synthetic data:} 
Learning from synthetic data avoids tedious manual \xiao{labeling} of real images. \xiao{It also provides} a platform to study high-level image understanding tasks without having to wait for low-level recognition problems to be solved. Moreover, synthetic data can be collected in large amounts and with high density, allowing us to learn rich models. 
Previous works have looked at learning recognition models from synthetic data. For instance, computer graphics models were used to synthesize data to learn human pose~\cite{Shotton_2013} and chair models~\cite{Aubry_2014}. Clipart data has been used to learn models of fine-grained interactions between people~\cite{Antol_2014}. \cite{Lim_2011} warps images of one category to use them as examples for other categories. \cite{Kaneva_2011} uses synthetic images to evaluate low-level image features. Human-\xiao{created} clipart images have been used to learn which semantic features (occurrence or co-occurrence of objects, pose, expression, relative location, \etc) are relevant to the meaning of a scene~\cite{Zitnick_2013_CVPR} and to learn spatio-temporal common sense to model scene dynamics~\cite{Fouhey_2014}.
\xiao{In this work, we learn our models from human-created clipart scenes. We also use clipart to ``imagine'' scenes in order to solve the FITB and VP tasks. Though the abstract scenes \cite{Zitnick_2013_CVPR} are not photo-realistic, they offer, more importantly, a semantically rich world} where one can effectively generate scenes and learn semantic variations of sentences and scenes, free from the bottlenecks of (still) imperfect object recognition and detection. Despite being synthetic, it has been shown that semantic concepts learnt from abstract scenes can generalize to real images~\cite{Antol_2014}.

\section{Dataset}
\label{sec:dataset}


We build our FITB and VP datasets on top of the Abstract Scenes Dataset\footnote{\url{http://research.microsoft.com/en-us/um/people/larryz/clipart/abstract_scenes.html}}, which has 10,020 human-created abstract scenes of a boy and a girl playing in the park. The dataset contains 58 clipart objects including the boy (Mike), the girl (Jenny), toys, background objects like trees and clouds, animals like dogs and cats, food items like burgers and pizzas, \etc. A subset of these objects are placed in the scene at a particular location, scale, and orientation (facing left or right). The boy and the girl can have different poses (7) and expressions (5). 
Each one of the 10,020 scenes 
has textual descriptions written by two different people. We use this clipart as the representation within which we will ``imagine'' our scenes. We also use this dataset to learn visual common sense. 
While more clipart objects, expressions, poses, \etc. can enable us to learn more comprehensive visual common sense, this dataset has been shown to contain semantically rich information~\cite{Zitnick_2013_CVPR,Zitnick_2013_ICCV}, sufficient to begin exploring our proposed tasks. We now describe our approach to creating our FITB and VP datasets.

\subsection{\xiao{Fill-in-the-blank (FITB)} Dataset}

\xiao{Every description in the Abstract Scenes Dataset consists of three short sentences, typically describing different aspects of the scene while also forming a coherent description. Since we have two such descriptions for every scene, we arbitrarily place one of the two descriptions (for all scenes) into the source set and the other into the distractor set.} 
For each image, we randomly drop one sentence from its source description to form an FITB question. We group this dropped sentence with \xiao{3 random sentences} from descriptions of other images in the distractor set. The \xiao{FITB} task is to correctly identify which sentence in the options belongs to the original description in the question.

Removing questions where the NLP parser produced degenerate outputs, our resulting FITB dataset contains 8,959 FITB questions -- 7,198 for training and 1,761 for testing. Figure~\ref{fig:scene_gen_fitb} shows one example FITB question from our dataset. The scenes corresponding to the questions in the training set are available for learning visual common sense and text-image correspondence. The scenes corresponding to the test questions are not available at test time.

FITB is a challenging task. Many scenes share the same visual elements such as Mike and Jenny playing football. Sometimes the distractor options may seem just as valid as the ground truth option, even to humans. We conduct studies on human performance on the test set. We had 10 different subjects on Amazon Mechanical Turk (AMT) answer the FITB questions. To closely mimic the task given to machines, subjects were not shown the corresponding image. We found that the majority vote response (\ie mode of responses) across 10 subjects agreed with the ground truth 52.87\% of the time (compared to random guessing at 25\%).

Some questions have disagreements among the subjects, while other questions have consistent responses across subjects. We find that 41\% of the questions in our dataset have 7 or more subjects agreeing on the response. Of these questions, the mode of the responses across subjects agrees with the ground truth 69\% of the time. Interestingly, on the remaining 31\% of the questions, 7 out of 10 subjects agree on the \emph{wrong} response. In our experiments, we report accuracies relative to the ground truth response, as well as relative to the response that most subjects agree on (the latter might be more relevant from an AI perspective -- if the goal is to produce human-like responses).

\begin{figure}[t]
\begin{center}
   \includegraphics[width=1\linewidth]{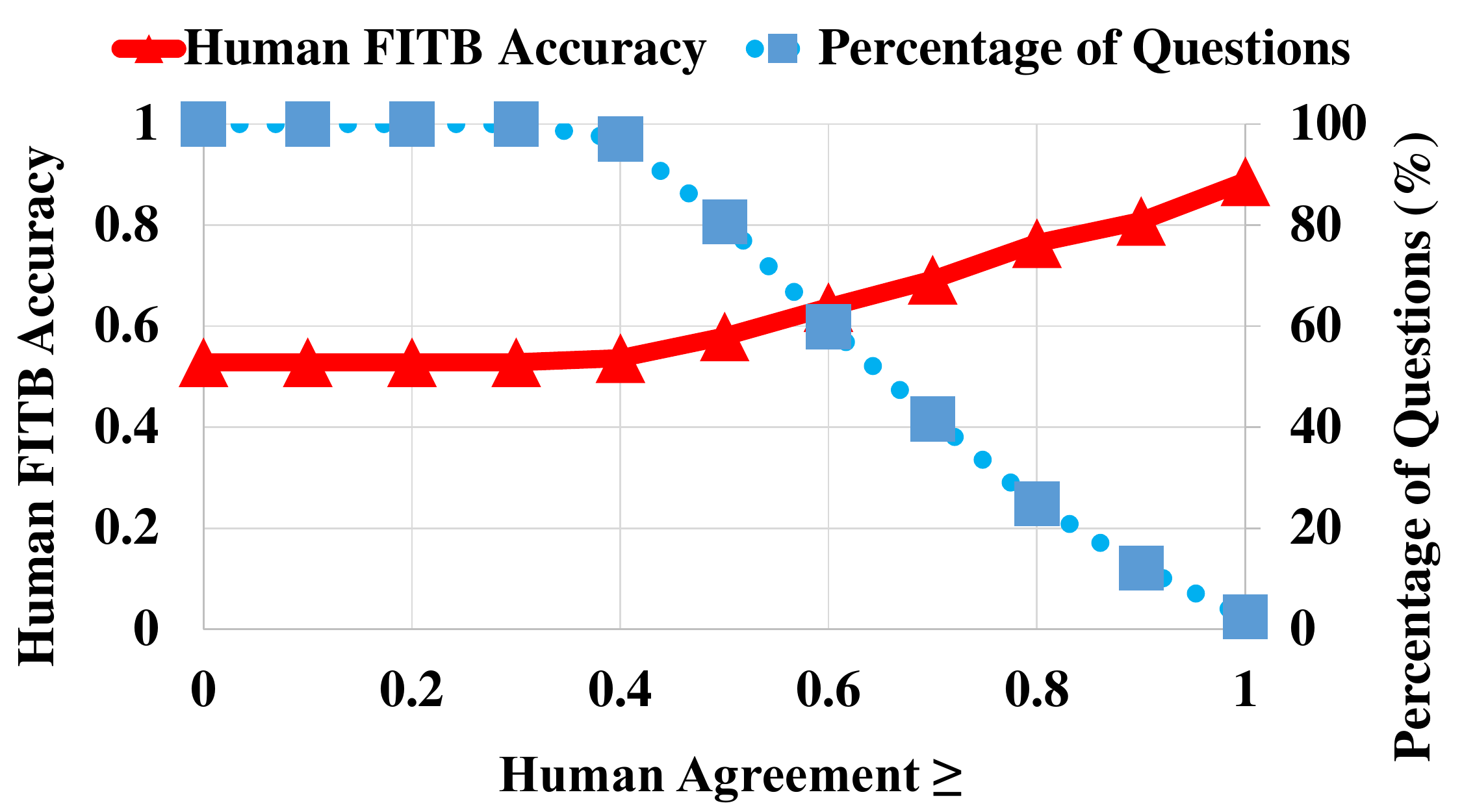}
\end{center}
\vspace{-10pt}
   \caption{Human performance vs. inter-human agreement on the FITB task. Mode of human responses is more accurate when subjects agree with each other.}
   \vspace{-10pt}
\label{fig:fitb_human_agreement}
\end{figure}

In Figure~\ref{fig:fitb_human_agreement}, we consider different subsets of the dataset formed by only considering questions where a certain minimum proportion of subjects agreed on the response (human agreement). For each subset, we can evaluate the accuracy of the mode response. We also look at what percentage of the dataset falls in each subset. Not surprisingly, human accuracy (mode agreeing with ground truth) correlates well with human agreement (percentage of subjects that agree with mode). Note that even if responses were random, \xiao{on average} 43\% of subjects would agree on the mode response. 


\subsection{\xiao{Visual Paraphrasing (VP)} Dataset}

The VP task is to tell if two descriptions are describing the same scene or two different scenes. The correct answer to a pair of descriptions written by two people describing the same scene is ``Yes'', while to randomly drawn descriptions from two different scenes is ``No''.

%

We build our VP dataset using all 10,020 scenes from the Abstract Scenes Dataset, resulting in a dataset with 10,020 positive pairs. We randomly sample 2 $\times$10,020 pairs as negatives. This leads to a total of 30,060 questions in our dataset. Of these, 24,000 are used for training and the rest 6,060 are used for testing. We choose the negative pairs separately in training and testing sets such that they do not overlap with each other. Figure~\ref{fig:scene_gen_vp} shows one example VP question from our dataset. 

We evaluate human performance on our test set. We had 10 different subjects on AMT solve our tasks.
We average their responses (0 for No and 1 for Yes) to obtain a score between 0 and 1 for each question. We can use this score to plot a precision-recall curve. Results show that humans can reliably solve this task with \xiao{94.78\%} average precision (AP), compared to chance at 33\%. 

\xiao{FITB and VP tasks are ways} to evaluate visual common sense. Some applications of FITB tasks may be automatic story telling and automatic Q\&A. Some applications of the VP task may be text-based image retrieval and generating multiple diverse descriptions of the same image.

\section{Approach}
\label{sec:approach}

We first (Section~\ref{sec:text}) describe the strong baseline approach of using textual features (common sense) to solve the FITB and VP tasks. We then describe our visual common sense model (Section~\ref{sec:visual}) and scene generation approach (Section~\ref{sec:scene_gen}). Finally in Section~\ref{sec:solve_tasks} we describe our approach to using our model to solve the FITB and VP tasks.

\subsection{Text Only Model}
\label{sec:text}


We first tokenize all words in our dataset and form a vocabulary (1,886 words for the FITB dataset and 2,495 for the VP dataset). We also form a vocabulary of pairs of words by selecting 100 pairs of words which have the highest mutual information in the training data and co-occur more than 100 times. 

Both FITB and VP involve reasoning about consistency between two descriptions (question and option for FITB and two input descriptions for VP). Given two descriptions $d_1$ and $d_2$, we extract three kinds of textual features from the pair. The first is term frequency\xiao{, commonly used for text classification and retrieval,} which counts how often each word from our vocabulary occurs in $(d_1,d_2)$ (both descriptions concatenated). The second is a 400D word co-occurrence vector indicating for each (of the 100) pair of words whether: 
(i) the first word occurred in $d_1$ and the second word occurred in $d_2$ or 
(ii) the first word occurred in $d_1$ and the second word did not occur in $d_2$ or 
(iii) the first word did not occur in $d_1$ and the second word occurred in $d_2$ or 
(iv) the first word did not occur in $d_1$ and the second word did not occur in $d_2$.
The third uses a \xiao{state-of-the-art} deep learning based word embedding representation learnt from a large text corpus. We use word2vec~\cite{Mikolov_2013} to represent each word with a \xiao{(default)} 200D vector. We then average the vector responses of all words in $(d_1,d_2)$. These features capture common sense knowledge about which words are used interchangeably to describe the same thing, which words tend to co-occur in descriptions, \etc.

\vspace{-10pt}
\paragraph{Fill-in-the-blank.}
For $N$ fill-in-the-blank questions and $M$ options per question, we denote the question body as $q_i, i \in \{1,\dots,N$\} and the options for $q_i$ as $o_{ij}, j \in \{1,\dots,M\}$. We denote the ground truth option for question $q_i$ as $o^{gt}_{i}$, and its index as $j^{gt}_{i}$.

The FITB problem is a ranking problem: given $q_i$, we wish to rank the correct option $o^{gt}_{i}$ above distractors $o_{ij}, j \ne j^{gt}_{i}$. For each question-option pair $(q_i, o_{ij})$, we extract the three kinds of textual features as described above using $d_1 = q_i$ and $d_2 = o_{ij}$. Concatenating these three gives us a 2,486D text feature vector $\phi^{text}_{fitb}(q_i, o_{ij})$. We compute scores $s_{ij}=w^{T}\phi^{text}_{fitb}(q_i, o_{ij})$ for each option that captures how likely $o_{ij}$ is to be the answer to $q_i$. We then pick the option with the highest score. We learn $w$ using a ranking SVM \cite{Chapelle_2010}:

\vspace{-15pt}
\begin{equation}
\begin{aligned}
\label{eq:rankSVM}
& \underset{w,\xi \ge 0}{\text{min}}
& & \frac{1}{2}\|w\|^2 + C \sum_{(i,j), j \ne j^{gt}} \xi_{ij}^2 \\
& \textit{s.t.}
& & w^{T}\phi^{text}_{fitb}(q_i, o^{gt}_{i})-w^{T}\phi^{text}_{fitb}(q_i, o_{ij}) \ge 1-\xi_{ij}, \\
&&& \forall (i,j), j \ne j^{gt}
\end{aligned}
\vspace{-5pt}
\end{equation}


%

\vspace{-10pt}
\paragraph{Visual paraphrasing.}
In visual paraphrasing, for each question $i$, the goal is to verify if the two given descriptions $q_{i1}$ and $q_{i2}$ describe the same image ($y_{i}=1$) or not ($y_{i}=-1$). We extract all three features described above using $d_1=q_{i1}$ and $d_2=q_{i2}$. Let's call this $\phi^{text}_{vp1}$. We extract the same features but using $d_1=q_{i2}$ and $d_2=q_{i1}$. Let's call this $\phi^{text}_{vp2}$. To ensure that the final feature representation is symmetric -- \ie $\phi^{text}_{vp}(q_{i1},q_{i2})=\phi^{text}_{vp}(q_{i2},q_{i1})$, we use $\phi^{text}_{vp} = [\phi^{text}_{vp1}+\phi^{text}_{vp2},|\phi^{text}_{vp1}-\phi^{text}_{vp2}| ]$ \ie a concatenation of the summation of $\phi^{text}_{vp1}$ and $\phi^{text}_{vp2}$ with the absolute difference between the two. This results in a $(2\times2,495)+(2\times200)+(2\times400)$ = 6,190D feature vector $\phi^{text}_{vp}$ describing $(q_{i1}, q_{i2})$. We then train a binary linear SVM to verify whether the two descriptions are describing the same image or not.





\subsection{Incorporating Visual Common Sense}
\label{sec:visual}



\xiao{Our model extends the baseline text-only model (Section~\ref{sec:text}) by using an ``imagined'' scene as an intermediate representation.
``Imagining'' a scene involves settings values for all of the variables (\eg presence of objects, their location) that are used to encode scenes. This encoding, along with priors within this abstraction that reason about which scenes are plausible, serve as our representation of visual common sense. }
This is in contrast with traditional knowledge base representations used to encode common sense via text~\cite{Zhu_2014,Pirsiavash_2014}. Exploring alternative representations of visual common sense is part of future work.

Given a textual description $S_i$, we generate a scene $I_i$. We first describe our scoring function \xiao{that} scores the plausibility of the $(S_i, I_i)$ pair. We then (Section~\ref{sec:scene_gen}) describe our scene generation approach. Our scoring function

\vspace{-20pt}
\begin{align}
\label{eq:main}
\Omega(I_i,S_i)=\Phi(S_i)+\Phi(I_i)+\Psi(I_i,S_i)
\end{align}

\vspace{-10pt}
captures textual common sense, \xiao{visual common sense and text-image correspondence}. The textual common sense term $\Phi(S_i) = w^{T}\phi^{text}(S_i)$ only depends on text and is the same as the text-only baseline model (Section~\ref{sec:text}). Of the two new terms, $\Phi(I_i)$  only depends on the scene and captures visual common sense -- it evaluates how plausible the scene is (Section~\ref{sec:visual}). Finally, $\Psi(I_i,S_i)$ depends on both the text description and the scene, and captures how consistent the imagined scene is to the text (Section~\ref{sec:joint}). We start by describing the representation we use to represent the description and to encode a scene via visual abstractions.



\vspace{-10pt}
\subsubsection{Scene and Description Encoding}
\label{sec:encoding}
\vspace{-5pt}
 

The set of clipart in our visual abstraction were described in Section~\ref{sec:dataset}. More details can be found in~\cite{Zitnick_2013_CVPR}. In the generated scenes, we represent an object $O_k$ using its presence $e_k\in\{0,1\}$, location ${x_k,y_k}$, depth $z_k$ (3 discrete scales), horizontal facing direction or orientation $d_k\in\{-1,1\}$ (left or right) and attributes $f_k$ (poses and expressions for the boy and girl).  The sentence descriptions $S_i$ are represented using a set of predicate tuples $T_l$ extracted using semantic roles analysis~\cite{Quirk_2012}. A tuple $T_l$ consists of a primary noun $A_l$, a relation $r_l$ and an optional secondary noun $B_l$. For example a tuple can be (Jenny, fly, Kite) or (Mike, be angry, N/A). 
There are 1,133 nouns and 2,379 relations in our datasets. Each primary noun $A_l$ and secondary noun $B_l$ is mapped to 1 of 58 objects $a_l$ and $b_l$ respectively which have the highest mutual information with it in training data. We found this to work reliably.


\vspace{-10pt}
\subsubsection{Visual Common Sense}
\vspace{-5pt}
\label{sec:visual}
We breakdown and introduce the factors in $\Phi(I_i)$ into per-object (unary) factors $\Phi^u(O_k)$ and between-object (pairwise) factors $\Phi^{pw}(O_{k_1},O_{k_2})$.

\vspace{-12pt}
\begin{align}
\Phi(I_i)=\sum_k{\Phi^u(O_k)}+\sum_{k_1,k_2}{\Phi^{pw}(O_{k_1},O_{k_2})}
\end{align}
\vspace{-5pt}

Per-object (unary) factors $\Phi^u(O_k)$ capture presence, location, depth, orientation and attributes. This scoring function will be parameterized by $w$'s\footnote{Overloaded notation with parameters learnt for the text-only baseline in Section~\ref{sec:text}} that are shared across all objects and pairs of objects. Let $L$ be the log probabilities (MLE counts) estimated from training data. For example, $L^{u}_e(e_k)=\log P(e_k)$, where $P(e_k)$ is the proportion of images in which object $O_k$ exists, and $L^{u}_{xyz}(x_k,y_k|z_k)= \log P(x_k,y_k|z_k)$, where $P(x_k,y_k|z_k)$ is the proportion of times object $O_k$ is at location $(x_k,y_k)$ given that $O_k$ is at depth $z_k$. 

\vspace{-17pt}
\begin{align}
\label{eq:visual_unary}
\nonumber \Phi^{u}(O_k)=& w^{u}_e L^{u}_e(e_k)+w^{u}_{xyz} L^{u}_{xyz}(x_k,y_k|z_k)+w^{u}_z L^{u}_z(z_k) \\
& +w^{u}_d L^{u}_d(d_k)+w^{u}_f L^{u}_f(f_k)
\end{align}

Between-object (pairwise) factors $\Phi^{pw}(O_{k_1},O_{k_2})$ capture co-occurrence of objects and their attributes, as well as relative location, depth and orientation.

\vspace{-17pt}
\begin{align}
\label{eq:visual_pw}
\nonumber \Phi^{pw}(O_{k_1},& O_{k_2})=w^{pw}_e L^{pw}_e(e_{k_1},e_{k_2})+w^{pw}_{xyd} L^{pw}_{xyd}(dx,dy) \\
\nonumber &+w^{pw}_z L^{pw}_z(z_{k_1},z_{k_2})+w^{pw}_d L^{pw}_d(d_{k_1},d_{k_2}) \\
&+w^{pw}_f L^{pw}_f(f_{k_1},f_{k_2})
\end{align}

Here the relative x-location is relative to the orientation of the first object \ie $dx=d_{k_1}(x_{k_1}-x_{k_2})$. Relative y-location is $dy=y_{k_1}-y_{k_2}$. These capture where $O_{k_2}$ is from the perspective of $O_{k_1}$. The space of $(x,y,z)$ is quite large (typical image size is \xiao{500 x 400}). So to estimate the probabilities reliably, we model the locations with GMMs. In particular, the factor $L^{u}_{xyz}(x_k,y_k|z_k)$ is over \xiao{27} GMM components and $L^{pw}_{xyd}(dx,dy)$ is over \xiao{24} GMM components.

Notice that since the parameters are shared across all objects and pairs of objects, so far we have introduced 5 parameters in Equation~\ref{eq:visual_unary} and 5 parameters in Equation~\ref{eq:visual_pw}. The corresponding 10 \xiao{log-likelihood} terms can be thought of as features representing visual common sense. The parameters will be learnt to optimize for the FITB (\xiao{ranking} SVM) or VP (binary SVM) tasks similar to the text-only baseline described in Section~\ref{sec:text}.

\vspace{-10pt}
\subsubsection{Text-Image Consistency}
\label{sec:joint}
\vspace{-5pt}

We now \xiao{discuss terms in our model that} score the consistency between an imaged scene and a textual description. We breakdown and introduce the \xiao{text-image correspondence} factors in $\Psi(I_i,S_i)$ in Equation~\ref{eq:main} into per-noun factors $\Psi^{n+}(I_i,T_l)$ and per-relation factors $\Psi^{r+}(I_i,T_l)$ for objects that are mentioned in the description, and default per-object factors $\Psi^{u-}(O_k)$ and default between-object factors \xiao{$\Psi^{pw-}(O_{k_1},O_{k_2})$} when the respective objects are not mentioned in the description.

\vspace{-12pt}
\begin{align}
\nonumber \Psi(I_i,S_i)=&\sum_l{\Psi^{n+}(I_i,T_l)}+\sum_{l}{\Psi^{r+}(I_i,T_l)}\\
& +\sum_{k\not\in S_i}{\Psi^{u-}(O_k)}+\sum_{k_1,k_2\not\in S_i}{\Psi^{pw-}(O_{k_1},O_{k_2})}
\end{align}
\vspace{-10pt}

The per-noun factors $\Psi^{n+}(I_i,T_l)$ capture object presence conditioned on the nouns (both primary and secondary) in the tuple, and object attributes conditioned on the nouns as well as relations in the tuple. For instance, if the tuple $T_l$ is ``(Jenny, kicks, ball)'', these terms reason about the likelihood that Jenny and ball exist in the scene, that Jenny has a certain attribute (\eg kicking pose), \etc. Again, the likelihood of each concept is scored by its log probability in the training data.

\vspace{-12pt}
\begin{align}
\nonumber \Psi^{n+}(I_i,&T_l)= w^{n+}_{abe} \big( L^{n+}_e(e_{a_l}|a_l) +  L^{n+}_e(e_{b_l}|b_l) \big) \\
&+ w^{n+}_{arf} L^{n+}_{arf}(f_{a_l}|a_l,r_l) + w^{n+}_{brf} L^{n+}_{brf}(f_{b_l}|b_l,r_l)
\end{align}
\vspace{-10pt}

The per-relation factors $\Psi^{r+}(I_i,T_l)$ capture relative object location (where is $b_l$ relative to $a_l$ and vice versa), depth and orientation conditioned on the relation. Note that these factors are shared across all objects because ``wearing'' in (Mike, \xiao{wears}, hat) and (bear, \xiao{wears}, crown) is expected to have similar visual instantiations.

\vspace{-14pt}
\begin{align}
\nonumber \Psi^{r+}(I_i,&T_l)=w^{r+}_{rxyd} L^{r+}_{rxyd}(dx,dy|r_l) \\
\nonumber &+ w^{r+}_{rxyd'} L^{r+}_{rxyd'}(dx',dy'|r_l) \\
&  + w^{r+}_{rz} L^{r+}_{rz}(z_{a_l},z_{b_l}|r_l) + w^{r+}_{rd} L^{r+}_{rd}(d_{a_l},d_{b_l}|r_l)
\end{align}
\vspace{-11pt}

Here $dx'=d_{b_l}(x_{b_l}-x_{a_l})$ and $dy'=y_{b_l}-y_{a_l}$ captures where the primary object is relative to the secondary object.

The default per-object factors $\Psi^{u-}(O_k)$ and the default between-object factors $\Psi^{pw-}(O_{k_1},O_{k_2})$ capture default \xiao{statistics} when an object or a pair of objects is not mentioned in the description. $\Psi^{u-}(O_k)$ captures the default presence and attribute whereas $\Psi^{pw-}(O_{k_1},O_{k_2})$ captures the default relative location, depth and orientation.

The default factors are object-specific since each object has a different prior depending on its semantic role in scenes.
The default factors capture object states conditioned on the object not being mentioned in a description. We use notation $D$ instead of $L$ to stress this point. For example $D^{u-}_e(e_k|S_i)=\log P(e_k|k \not\in S_i)$, $D^{pw-}_z(z_{k_1},z_{k_2}|S_i)=\log P(z_{k_1},z_{k_2}|k_1,k_2 \not\in S_i)$.

\vspace{-14pt}
\begin{align}
\nonumber & \Psi^{u-}(O_k)=w^{u-}_{abe} D^{u-}_{abe}(e_k|S_i) + w^{u-}_{abrf} D^{u-}_{abrf}(f_k|S_i) \\
\nonumber & \Psi^{pw-}(O_{k_1},O_{k_2})=w^{pw-}_{rxyd} D^{pw-}_{rxyd}(dx,dy|S_i) \\
 &+ w^{pw-}_{rz} D^{pw-}_{rz}(z_{k_1},z_{k_2}|S_i) + w^{pw-}_{rd} D^{pw-}_{rd}(d_{k_1},d_{k_2}|S_i) 
\end{align}
\vspace{-11pt}

We have now introduced an additional 12 $w$ parameters (total 22) that are to be learnt to solve the FITB and VP tasks. Notice that this is in stark contrast with the thousands of parameters we learn for the text-only baseline (Section~\ref{sec:text}).


\begin{figure}[t]
\begin{center}
   \includegraphics[width=1\linewidth]{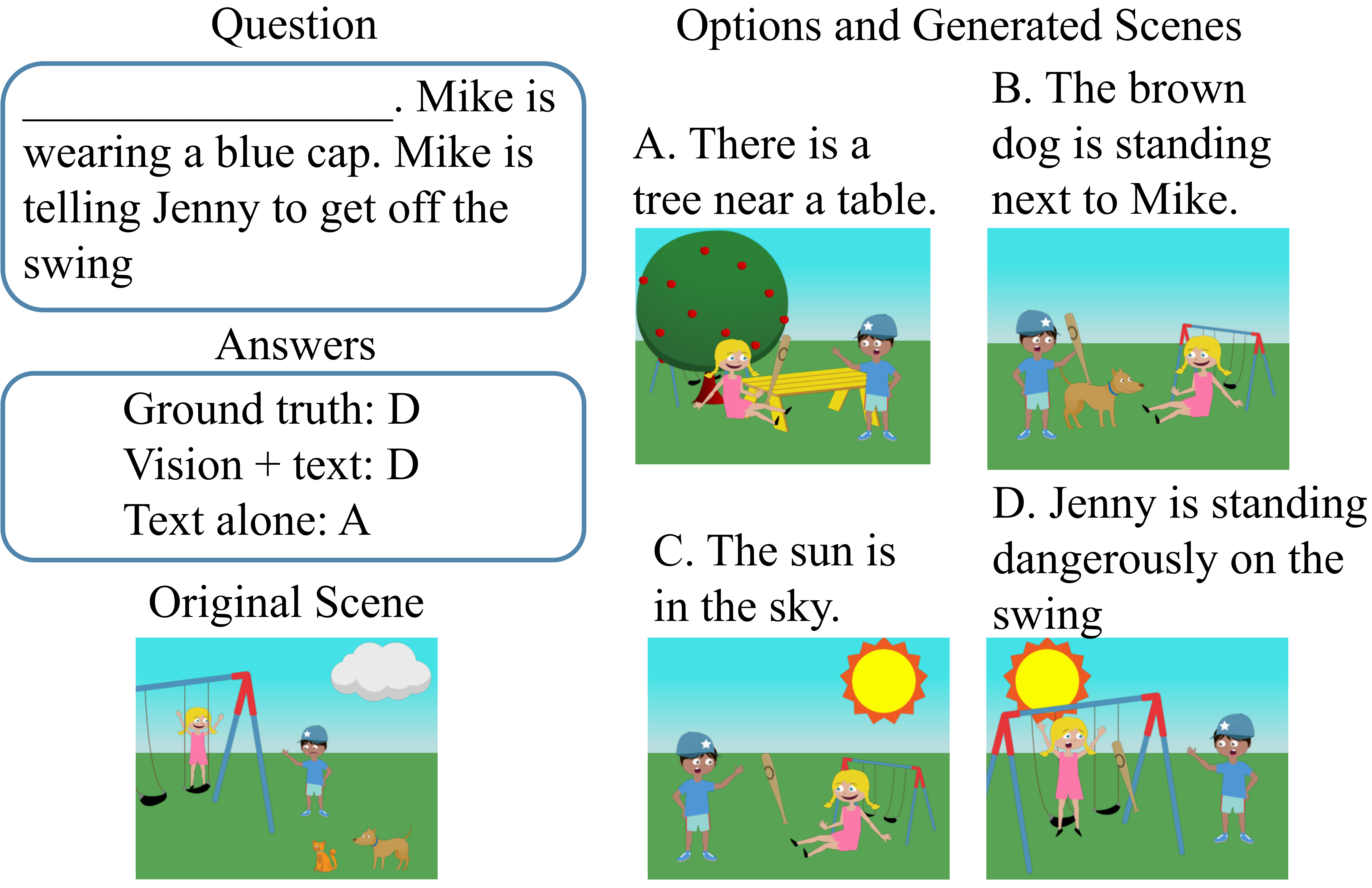}
\end{center}
\vspace{-10pt}
   \caption{Scenes generated for an example FITB question.}
   \vspace{-10pt}
\label{fig:scene_gen_fitb}
\end{figure} 

\subsection{Scene Generation}
\label{sec:scene_gen}
Given an input description, we extract tuples as described earlier in Section~\ref{sec:encoding}. We then use the approach of Zitnick~\etal~\cite{Zitnick_2013_ICCV} to generate a scene corresponding to the tuples. Briefly, it sets up a Conditional Random Field (CRF) model with a scoring function very similar to $\Phi(I_i)+\Psi(I_i,S_i)$. It samples scenes from this model using Iterative Conditional Modes with different initializations. Details can be found in~\cite{Zitnick_2013_ICCV}.

\subsection{Answering Questions with Imagined Scenes}
\label{sec:solve_tasks}

\textbf{Fill-in-the-blank. }
For FITB, we generate one scene using each question-answer pair $S_{ij}=(q_i,o_{ij})$. Fig. \ref{fig:scene_gen_fitb} shows qualitative examples of scenes generated for FITB. From the question-answer pair $S_{ij}$ and the generated scenes $I_{ij}$, we extract features corresponding to our scoring function (Equation~\ref{eq:main}) and use them to learn the \xiao{ranking} SVM (Equation~\ref{eq:rankSVM}) to answer FITB questions. We choose the ranking SVM C \xiao{parameter} using 5 fold cross validation.

\begin{figure}[t]
\begin{center}
   \includegraphics[width=1\linewidth]{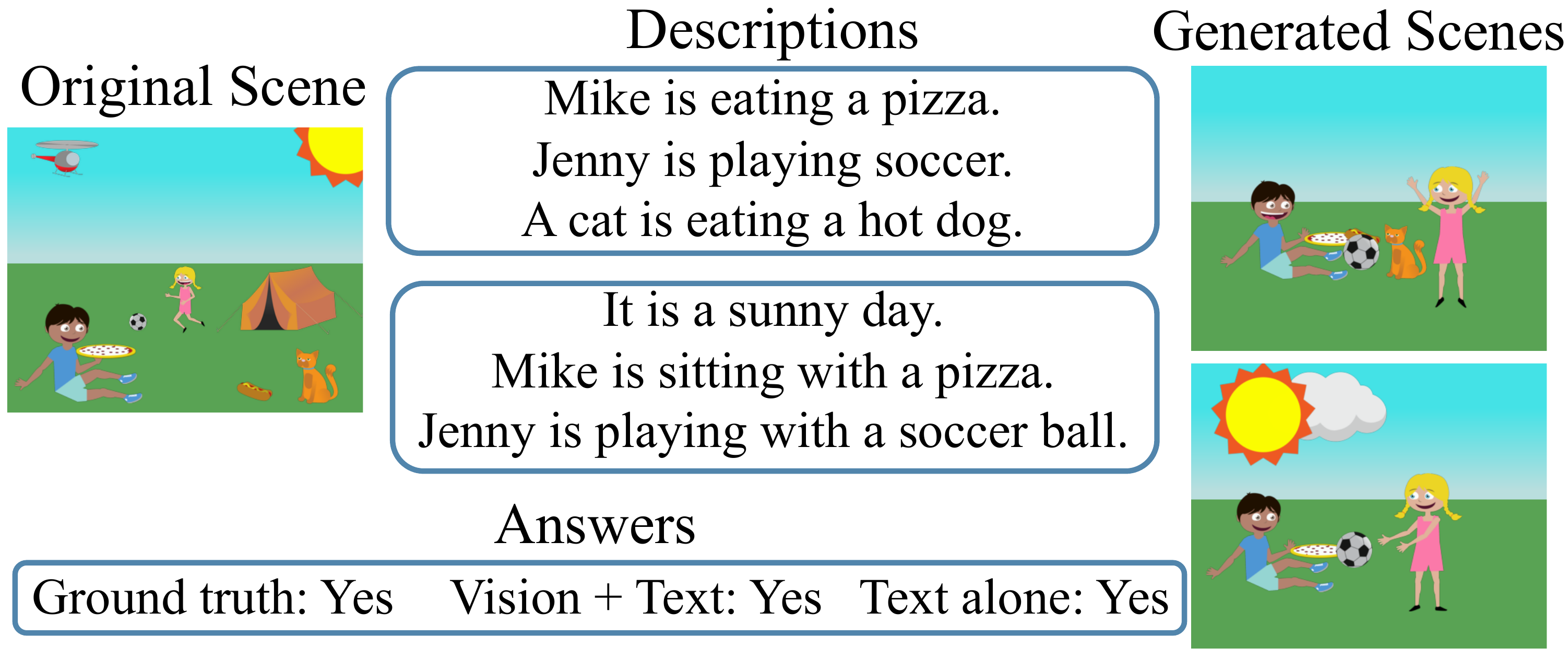}
\end{center}
\vspace{-10pt}
   \caption{Scenes generated for an example VP question.}
   \vspace{-10pt}
\label{fig:scene_gen_vp}
\end{figure}

\textbf{Visual paraphrasing. }
For VP we generate one scene for each description $S_{i1}=q_{i1}$ and $S_{i2}=q_{i2}$ in the input pair of descriptions. Fig. \ref{fig:scene_gen_vp} shows qualitative examples of scenes generated for VP. We capture the difference between the two sentence descriptions by pairing the generated scenes with the \emph{other} description \ie we compute $\Omega(I_{i1},S_{i2})$ and $\Omega(I_{i2},S_{i1})$ (Equation~\ref{eq:main}). We extract features for both combinations, concatenate the addition of the features and the absolute difference of the features to make the mapping symmetric. These features are used to train a binary SVM that determines whether the input pair of descriptions are describing the same scene or not. We choose the SVM C \xiao{parameter} using 5 fold cross validation.

\section{Experiments and Results}
\label{sec:result}

\subsection{Fill-in-the-blank}



We present results of our approach on the FITB dataset in Table \ref{table:fitb_result}. Our approach of ``imagining'' and joint visual-text reasoning achieves 48.04\% accuracy, significantly outperforming the text-only baseline (44.97\%) by 3.07\% using only 22 extra feature dimensions (compared to 2,486 dimensions of the baseline). This brings the performance closer to human performance at 52.87\%. Leveraging visual common sense does help answering these seemingly purely text-based questions.

By breaking down our 22 parameters (corresponding to visual features) into object presence ($w^{u}_e$, $w^{pw}_e$, $w^{n+}_{abe}$, $w^{u-}_{abe}$, 4D), attribute ($w^{u}_f$, $w^{pw}_f$, $w^{n+}_{arf}$, $w^{n+}_{brf}$, $w^{u-}_{abrf}$, 5D) and spatial configuration ($w^{u}_{xyz}$, $w^{u}_z$, $w^{u}_d$, $w^{pw}_{xyd}$, $w^{pw}_z$, $w^{pw}_d$, $w^{r+}_{rxyd}$, $w^{r+}_{rxyd'}$, $w^{r+}_{rz}$, $w^{r+}_{rd}$, $w^{pw-}_{rxyd}$, $w^{pw-}_{rz}$, $w^{pw-}_{rd}$, 13D) categories, we study their individual contribution to FITB performance on top of the text baseline. Object presence contributes the most (47.02\%), followed by attribute (46.39\%), while spatial information does not help (44.80\%). In fact, only using presence and attribute features achieves 48.60\%, slightly higher than using all three (including spatial). Visual features \xiao{alone} perform poorly (33.67\%), which is expected given the textual nature of the task. But \xiao{they} clearly provide useful complementary information over text. In fact, text-alone (baseline), vision+text (our approach) and humans all seem to make complementary errors. Between text-alone and vision+text, 54.68\% of the questions are correctly answered by at least one of them. And between text-alone, vision+text and human, 75.92\% of the questions are correctly answered.

\begin{table}[t]
\begin{center}
\begin{tabular}{l*{1}{c}r}
Approach&  Fill-in-the-blank  \\
        &  Accuracy(\%)   \\
\hline
Random & 25.00 \\
Text baseline & 44.97 \\
Visual & \xiao{33.67} \\
Text + visual (presence) & 47.02 \\
Text + visual (attribute) & 46.39 \\
Text + visual (spatial) & 44.80 \\
Text + visual (presence,attribute) & \textbf{48.60}  \\
Text + visual (all) & 48.04    \\
\hline
Human Mode & 52.87    \\
\end{tabular}
\end{center}
\vspace{-10pt}
\caption{Fill-in-the-blank performance of different approaches. 
}
\vspace{-10pt}
\label{table:fitb_result}
\end{table}

Our model is capable of imagining scenes that may contain more objects than the ones mentioned in text. 
Our model when using only presence does 47.02\%, while a visual common sense agnostic model that only infers objects mentioned in the tuples ($a_l$ and $b_l$) does 46.62\%. This further demonstrates the need for visual common sense based imagination, and not treating the text at face value.

In addition to predicting ground truth, we also study how well our approach can mimic \xiao{human responses}. Our approach matches the human majority vote (mode) response 39.35\% of the times (text alone: 36.40\%). When re-trained using the human mode as the labels, the performance increases to 45.43\%. The text-only baseline method does 42.25\%. These results suggest that mimicking human is a more challenging task (text-only was at 44.97\% when training on and predicting ground truth). Note that visual common sense is also useful when mimicking humans. 

\begin{figure}
\begin{center}
   \includegraphics[width=1\linewidth]{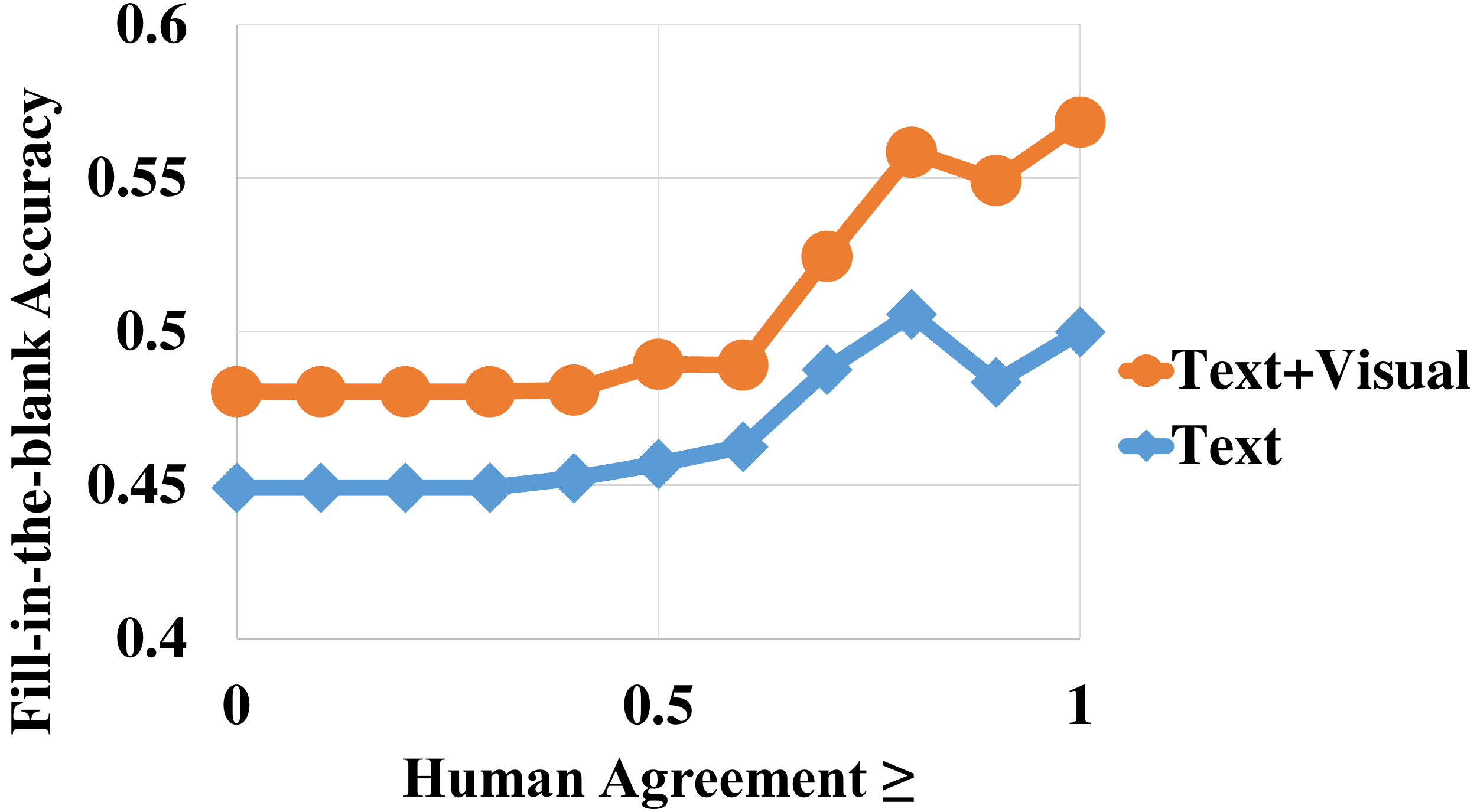}
\end{center}
\vspace{-10pt}
   \caption{FITB performance on subsets of the test data with varying amounts of human agreement. The margin of improvement of our approach over the baseline increases from 3\% on all questions to 6\% on  questions with high human agreement.}
   \vspace{-10pt}
\label{fig:fitb_subset}
\end{figure}

We also study how the performance of our approach varies based on the difficulty of the questions. We consider questions to be easy if humans agree on the response. We report performance of the text baseline and our model on subsets of the FITB test set where at least $K$ people agreed with the mode. Fig. \ref{fig:fitb_subset} shows performance as we vary $K$. On questions with higher human agreement, the visual approach outperforms the baseline by a larger margin. Qualitative results can be found in the supplementary material.

\subsection{Visual Paraphrasing}




\begin{table}
\begin{center}
\begin{tabular}{l*{1}{c}r}
Approach&  Visual Paraphrasing   \\
        &  Average Precision(\%)   \\
\hline
Random & 33.33 \\
Text baseline & 94.15 \\
Visual & \xiao{91.25} \\
Text + visual (presence) & 95.08 \\
Text + visual (attribute) & 94.54 \\
Text + visual (spatial) & 94.75 \\
Text + visual (presence,attribute) & 95.47  \\
Text + visual (all) & \textbf{95.55}    \\
\hline
Human Average & 94.78\\
\end{tabular}
\end{center}
\vspace{-10pt}
\caption{Visual paraphrasing performance of different approaches. 
}
\vspace{-10pt}
\label{table:vp_result}
\end{table}

We present results of our approach on the VP dataset in Table. \ref{table:vp_result}. Our approach of generating and reasoning with scenes does 1.4\% better than reasoning only with text. In this task, the performance of the text-based approach is already close to human, while vision pushes it even further to above human performance\footnote{Likely due to noise on MTurk.}.

Similar to the \xiao{FITB} task, we break down the contribution of visual features into object presence, attribute and spatial configuration categories. Presence shows the most contribution (0.93\%). Spatial configuration features also help (by 0.60\%) in contrast to FITB. See Table~\ref{table:vp_result}.

In VP, a naive scene generation model that only imagines objects that are mentioned in the description does 95.01\% which is close to 95.08\% where extra objects are inferred. We hypothesize that the VP task is qualitatively different from FITB. In VP, important objects that are relevant to semantic distance between sentences tend to be mentioned in the sentences. What remains is to reason about the attributes and spatial configurations of the objects. In FITB, on the other hand, inferring the unwritten objects is critical to identify the best way to complete the description. The VP task can be made more challenging by sampling pairs of descriptions that describe semantically similar scenes. In fact, the Abstract Scenes dataset contains groups of semantically scenes~\cite{Zitnick_2013_CVPR}. Exploring this is part of future work. Some qualitative results can be found in the supplementary material.
 
We would like to stress that FITB and VP are purely textual tasks as far as the input modality is concerned. The visual cues that we incorporate are entirely ``imagined''. Our results clearly demonstrate that a machine that imagines and uses visual common sense performs better at these tasks than a machine that does not.



\section{Discussion}
\label{sec:discussion}

\xiao{Leveraging} visual knowledge to solve non-visual tasks may seem counter-intuitive. Indeed, with sufficient training data, one may be able to learn a sufficiently rich text-based model.  However in practice, good intermediate representations provide benefits. This is the role that parts and attributes have played in recognition~\cite{Lampert_2009,Felzenszwalb_2010,Zhang_2013}. 
In this work, the imagined scenes form this intermediate representation that allows us to encode visual common sense. 

In this work, we choose clipart scenes as our modality to ``imagine'' the scene and harness the power of visual common sense. This is analogous to works on physical reasoning that use physics to simulate physical processes~\cite{Hamrick_2011}. These are both qualitatively different from traditional knowledge bases~\cite{Chen_2013,Zhu_2014}, where relations between instances are explicitly represented and used during inference. Humans cannot always verbalize their reasoning process. Hence, using non-explicit representations of common sense has some appeal. Of course, alternate approaches, including more explicit representations of visual common sense are worth investigating.


Improved scene generation models that better translate from text to vision, and better features and modalities to use the generated scenes to answer non-visual questions, could also show improvements. In our experiments we already show that a better scene generation model that infers objects beyond what the text mentions shows better performance. Instead of generating one image per text description, one could consider generating multiple diverse images to better capture the underlying distribution~\cite{Batra_2012}. With more visual data, one can also expect to learn more sophisticated joint text-image representations. Our scoring function is akin to a Conditional Random Field model, similar to the scene generation model~\cite{Zitnick_2013_ICCV}. One could envision learning the scene generation model and visual common sense models jointly, \ie learning to infer scenes for the FITB or VP tasks. The generated scenes capture a semantically rich space. It would be interesting to study other tasks that can benefit form this intermediate representation.

\section{Acknowledgment}
This work was supported in part by a Google Faculty Research Award and The Paul G. Allen Family Foundation Allen Distinguished Investigator award to Devi Parikh. We thank Larry Zitnick for helpful discussions and his code.

{\small
\bibliographystyle{ieee}
\bibliography{egpaper_for_review}
}


\section*{Supplementary material}

# Don't Just Listen, Use Your Imagination: Leveraging Visual Common Sense for Non-Visual Tasks

Supplemental Material

# Coarse and Fine-grained Visual Paraphrasing

- The 10,020 scenes in the Abstract Scenes Dataset are generated from 1,002 sentences. For each of the 1,002 sentences 10 different people drew 10 scenes. And then a new set of workers described each of the 10 scenes (10,020 total).
- Scenes that are generated from the same sentence belong to the same semantic class, and therefore their sentence descriptions have similar semantic meanings.
- We study coarse-grained and fine-grained visual paraphrasing problems.
  - In the coarse-grained visual paraphrasing problem, the objective is to tell sentences describing one semantic class from another.
  - In the fine-grained visual paraphrasing problem, the objective is to tell sentences describing the same semantic class from each other.

# Coarse and Fine-grained Visual Paraphrasing

- In both coarse- and fine-grained settings, visual features show improvements on top of the text-only baseline.

| | Source of positive pairs of sentences | Source of negative pairs of sentences | Random | Text only | Text + Visual | Visual Improvement |
|---|---|---|---|---|---|---|
| Original (in main paper) | Same scene | Different scenes | 33.33% | 94.15% | 95.55% | +1.40% |
| Coarse-grained | Different scenes in the same semantic class | Scenes from different semantic classes | 33.33% | 84.19% | 86.15% | +1.96% |
| Fine-grained | Same scene | Different scenes in the same semantic class | 33.33% | 54.79% | 56.43% | +1.64% |

# Qualitative Results: Fill-in-the-blank

- Scenario 1: human, text baseline and our approach are all correct.

Question

Mike kicked the soocer ball.
_____.
The duck is afraid of the soccer ball

A. Jenny and mike are angry at the dog.

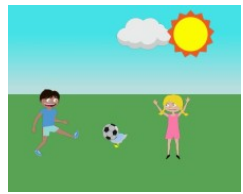

B. The bear has a hamburger and drink.

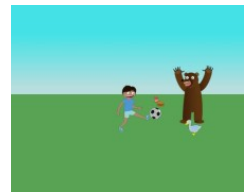

Answers

Ground Truth:  D

Human:  D (8/10)

Text baseline:  D

Vision + text:  D

Original Scene

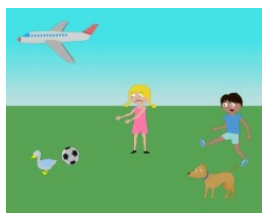

C. The grill is next to the tree.

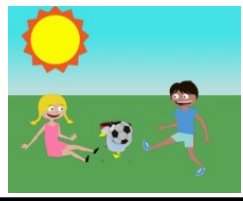

D. Jenny wants the soccer ball.

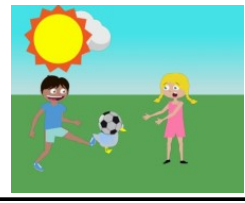

# Qualitative Results: Fill-in-the-blank

- Scenario 1: human, text baseline and our approach are all correct.

### Question

Jenny is standing on the swing.
Mike is feeling sad.
_____.

A. The dog is standing next to the table.

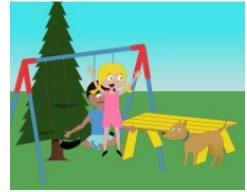

B. The sun is behind the tree.

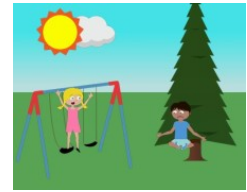

### Answers

Ground Truth: B

Human: B (5/10)

Text baseline: B

Vision + text: B

### Original Scene

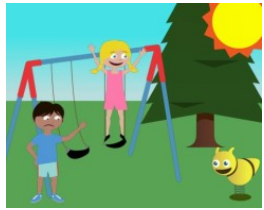

C. Jenny is angry because it is raining on her.

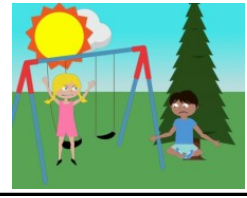

D. Jenny is near balloons.

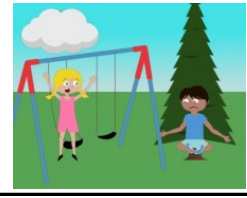

---

# Qualitative Results: Fill-in-the-blank

- Scenario 2: human and our approach are correct while text baseline is incorrect

### Question

_____.
Jenny is in the sandbox
The cat and Jenny have not left room for Mike

A. Mike sees a pie.

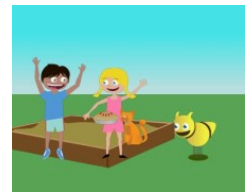

B. The cat is sitting next to Jenny.

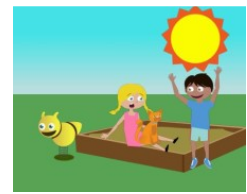

### Answers

Ground Truth: B

Human: B (9/10)

Text baseline: C

Vision + text: B

### Original Scene

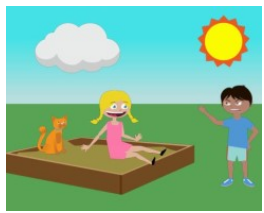

C. Mike and Jenny are sitting next a fire

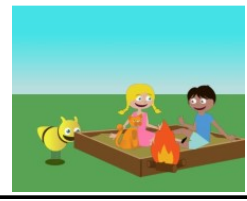

D. Jenny is playing in the sandbox.

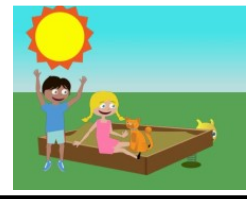

# Qualitative Results: Fill-in-the-blank

- Scenario 2: human and our approach are correct while text baseline is incorrect

**Question**

> Mike and Jenny are scared of the duck.
> Happy duck walks away.
> _____.

A. Mike was wearing his crown in the sandbox.

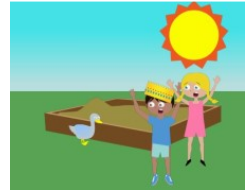

B. The ball hits the duck.

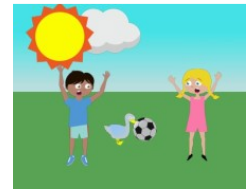

**Answers**

Ground Truth: B

Human: B (5/10)

Text baseline: A

Vision + text: B

**Original Scene**

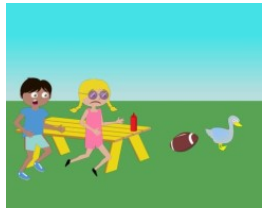

C. The sun is shining.

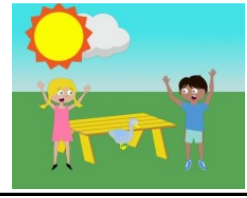

D. Mike is helping Jenny.

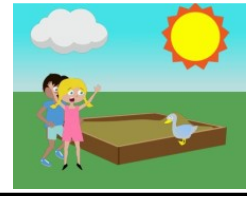

# Qualitative Results: Fill-in-the-blank

- Scenario 3: human and text baseline are correct while our approach is incorrect

**Question**

> Jenny is petting the cat.
> _____.
> No one is on the riding toy.

A. There is an apple tree behind Mike.

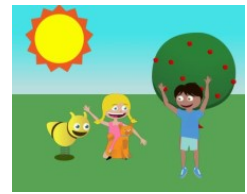

B. There are 3 hot dogs on the grill.

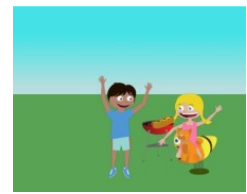

**Answers**

Ground Truth: C

Human: C (8/10)

Text baseline: C

Vision + text: A

**Original Scene**

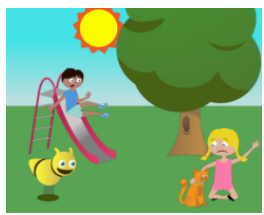

C. Mike is on the slide.

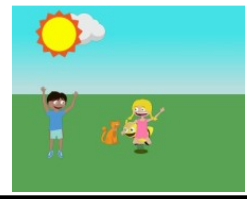

D. Jenny is happy to see Mike.

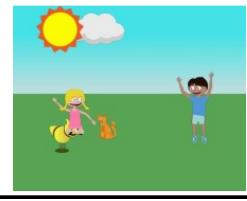

# Qualitative Results: Fill-in-the-blank

- Scenario 3: human and text baseline are correct while our approach is incorrect

**Question**

The burger is on the table.
_____.
Jenny is standing next to table.

A. Mike is flying a kite.

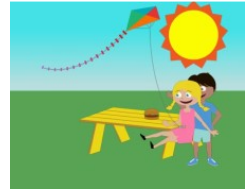

B. The dog is watching Jenny.

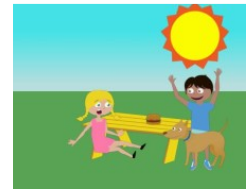

**Answers**

Ground Truth: D
Human: D (4/10)
Text baseline: D
Vision + text: B

**Original Scene**

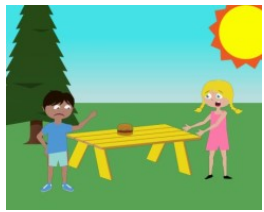

C. Jenny threw the frisbee.

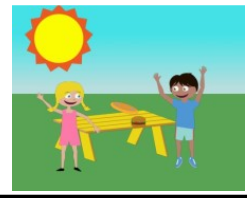

D. Mike is standing next to table.

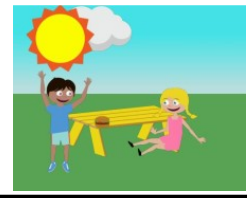

# Qualitative Results: Fill-in-the-blank

- Scenario 4: human is correct while text baseline and our approach are incorrect

**Question**

Jenny is holding a pink pail.
Mike threw the beach ball.
_____.

A. Mike is sitting next to the tree.

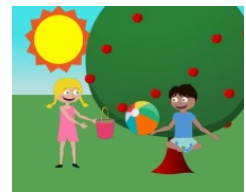

B. There are three hamburgers on the grill.

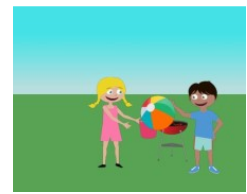

**Answers**

Ground Truth: D
Human: D (7/10)
Text baseline: C
Vision + text: A

**Original Scene**

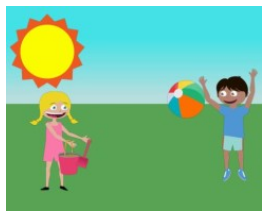

C. A rocket ship is flying in the sky.

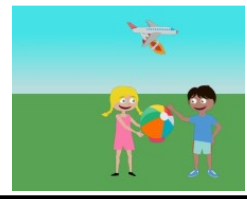

D. Jenny has a pink shovel.

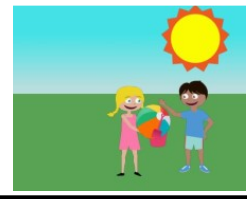

# Qualitative Results: Fill-in-the-blank

- Scenario 4: human is correct while text baseline and our approach are incorrect

**Question**

Jenny and Mike are fighting.
They are both wearing silly hats

A. Mike is holding a beach ball
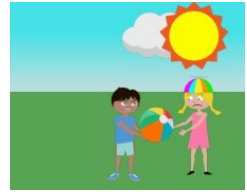

B. Mike is wearing the hat.
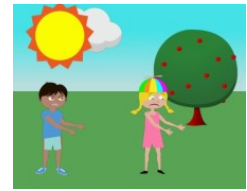

**Answers**

Ground Truth:   A

Human:   A (5/10)

Text baseline:   D

Vision + text:   D

**Original Scene**
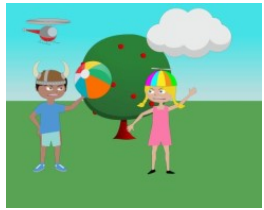

C. The dog is watching Mike.
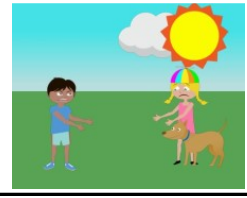

D. Jenny kicked the football.
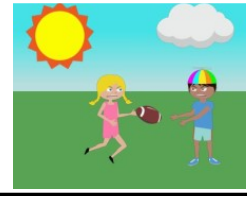

# Qualitative Results: Fill-in-the-blank

- Scenario 5: our approach and text baseline are correct while human is incorrect

**Question**

The duck is near the soccer ball.
Jenny is sitting near the slide.
_____.

A. Mike is standing under the hot air balloon
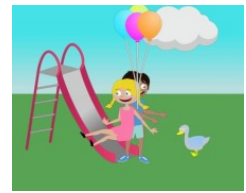

B. Mike is sitting next to the dog.
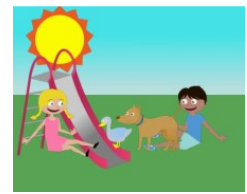

**Answers**

Ground Truth:   A

Human:   B (8/10)

Text baseline:   A

Vision + text:   A

**Original Scene**
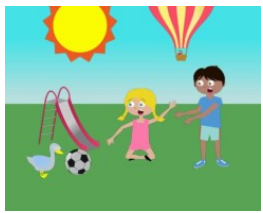

C. The snake is sliding behind Mike.
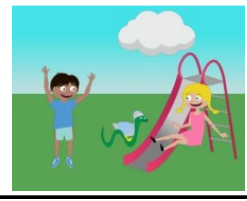

D. Mike is very surprised.
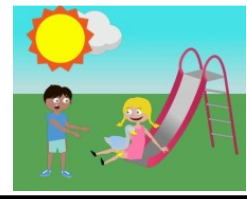

# Qualitative Results: Fill-in-the-blank

- Scenario 5: our approach and text baseline are correct while human is incorrect

**Question**

Mike is holding the ball.
_____.
Mike is playing with the cat.

A. Mike is wearing sun glasses.

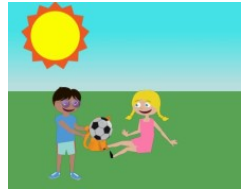

B. Jenny is sitting next to her juice.

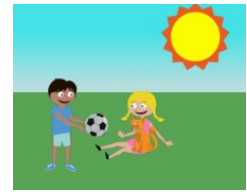

**Answers**

Ground Truth: A
Human: B (4/10)
Text baseline: A
Vision + text: A

**Original Scene**

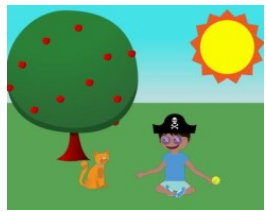

C. The bear is roaring angrily.

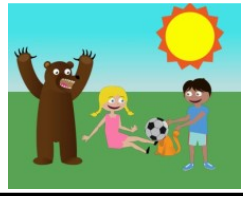

D. The duck is in the sandbox.

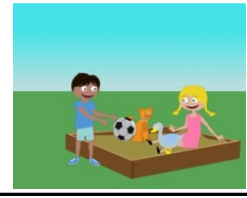

# Qualitative Results: Fill-in-the-blank

- Scenario 6: our approach is correct while human and text baseline are incorrect

**Question**

Mike is wearing a hat.
Jenny is holding the pizza.
_____.

A. Jenny is trying to catch the soccer ball

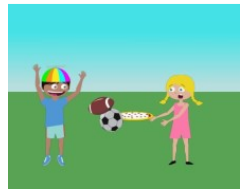

B. Mike is holding the shovel.

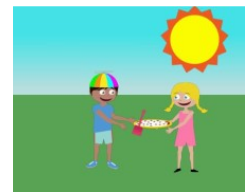

**Answers**

Ground Truth: D
Human: C (7/10)
Text baseline: B
Vision + text: D

**Original Scene**

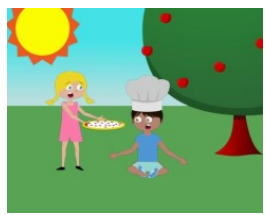

C. Mike and Jenny are happy.

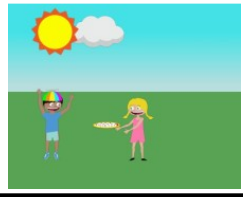

D. Mike is sitting on the grass.

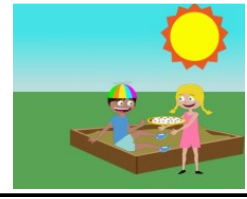

# Qualitative Results: Fill-in-the-blank

- Scenario 6: our approach is correct while human and text baseline are incorrect

### Question

_____.
Mike is sitting on the grass.
Jenny is standing by the table.

A. MIke is king for a day

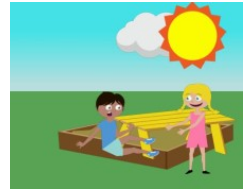

B. Jenny is angry at Mike.

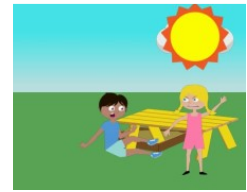

### Answers

Ground Truth: C
Human: D (5/10)
Text baseline: D
Vision + text: C

### Original Scene

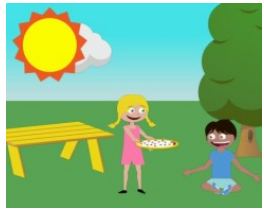

C. Jenny is holding a pizza.

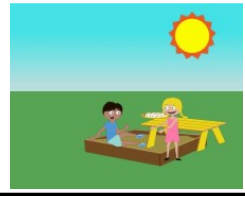

D. Mike is wearing a viking hat.

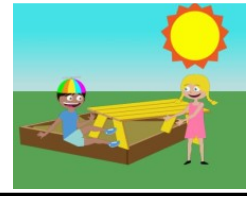

---

# Qualitative Results: Fill-in-the-blank

- Scenario 7: text baseline is correct while human and our approach are incorrect

### Question

_____.
Jenny is jumping up and down.
Mike is holding a frisbee.

A. Mike is wearing his viking hat.

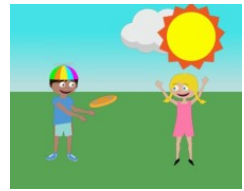

B. Mike and Jenny are camping

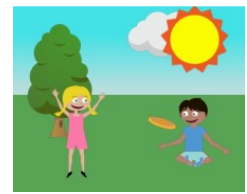

### Answers

Ground Truth: A
Human: B (7/10)
Text baseline: A
Vision + text: B

### Original Scene

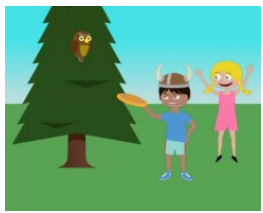

C. The rocket is soaring in the sky.

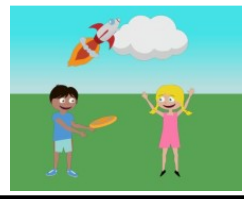

D. Jenny told the bear to leave.

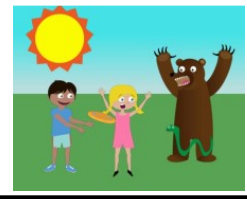

# Qualitative Results: Fill-in-the-blank

- Scenario 7: text baseline is correct while human and our approach are incorrect

### Question

_____.
Mike is playing in the sandbox.
Jenny wants to play with Mike.

A. Red apples grow on the tree.

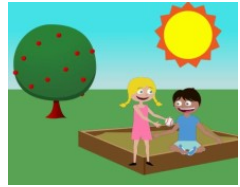

B. Mike is near jenny.

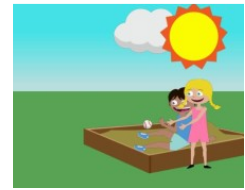

### Answers

| | |
|---|---|
| Ground Truth: | C |
| Human: | D (4/10) |
| Text baseline: | C |
| Vision + text: | D |

### Original Scene

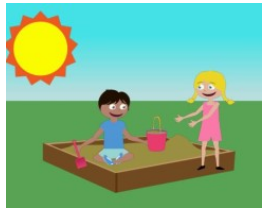

C. The sun is shining on Mike and Jenny.

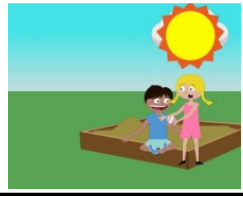

D. The pink shovel is on Jenny's lap.

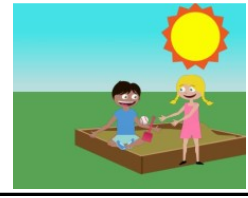

# Qualitative Results: Fill-in-the-blank

- Scenario 8: human, text baseline and our approach are all incorrect

### Question

Jenny is wearing a crown waving her hand.
_____.
The airplane is flying towards a giant cloud.

A. Mike is wearing a pirate hat.

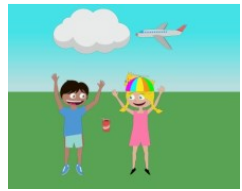

B. Mike is near the swings.

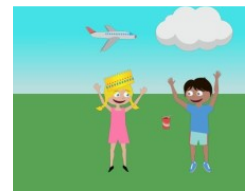

### Answers

| | |
|---|---|
| Ground Truth: | D |
| Human: | A (9/10) |
| Text baseline: | A |
| Vision + text: | A |

### Original Scene

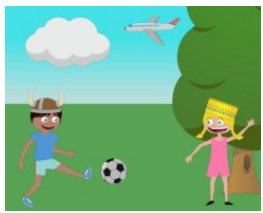

C. Mike has a baseball bat.

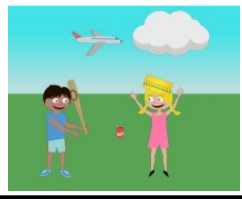

D. Mike is happily kicking the soccer ball.

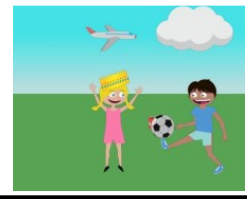

# Qualitative Results: Fill-in-the-blank

- Scenario 8: human, text baseline and our approach are all incorrect

### Question

> Jenny is upset she lost her balloons.
> Jenny is standing next to the cat.
> _____.

A. The airplane will not disturb them.

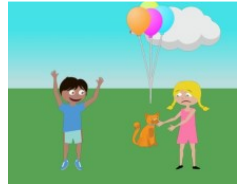

B. Mike is angry that the dog is not listening.

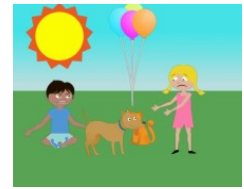

### Answers

Ground Truth: D

Human: C (4/10)

Text baseline: B

Vision + text: B

### Original Scene

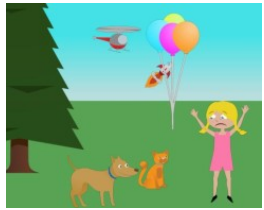

C. The cat is sitting by Jenny.

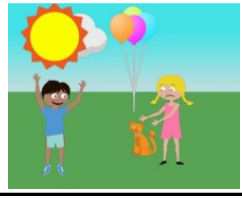

D. Jenny is afraid the rocket will hit the balloon.

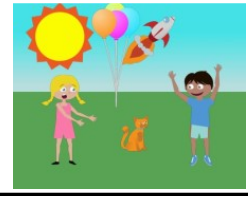

# Qualitative Results: Visual Paraphrasing

- Scenario 1: human, text baseline and our approach are all correct.

| Original Scene(s) | Descriptions | Generated Scenes | Answers |
|---|---|---|---|

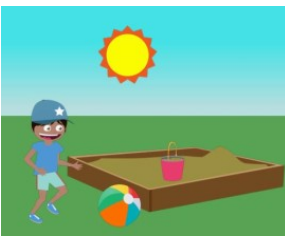

The bucket is in the sandbox. Mike runs to the ball. Mike is wearing a baseball cap.

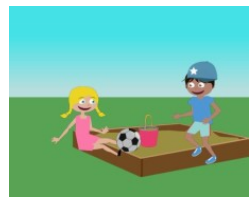

Ground truth
Yes

Human
1.3753

The bucket is in the sandbox. Mike runs to the ball. Mike is wearing a baseball cap.

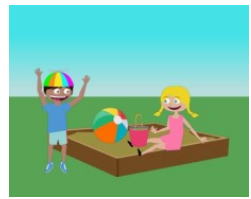

Text baseline
1.221

Vision + Text
2.0805

10

# Qualitative Results: Visual Paraphrasing

- Scenario 1: human, text baseline and our approach are all correct.

| Original Scene(s) | Descriptions | Generated Scenes | Answers |
|---|---|---|---|

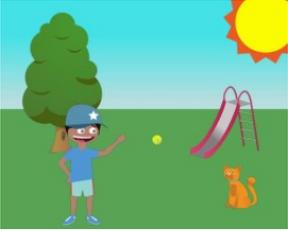

Mike loves throwing the tennis ball. There is a cat looking at Mike. Mike is playing with the cat.

Mike tries to play catch with the cat. The cat does not want to play catch. Mike threw the tennis ball to the cat.

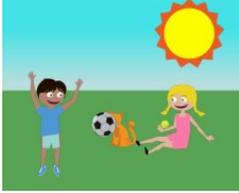

Ground truth
Yes

Human
4.2825

Text baseline
1.9647

Vision + Text
2.1077

---

# Qualitative Results: Visual Paraphrasing

- Scenario 1: human, text baseline and our approach are all correct.

| Original Scene(s) | Descriptions | Generated Scenes | Answers |
|---|---|---|---|

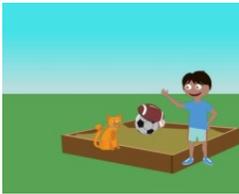

Mike is holding a hot dog Jenny is carring ketchup. Jenny is running.

Mike and Jenny are standing on the picnic table. Mike and Jenny are afraid of the bear. The owl is standing on the beach ball.

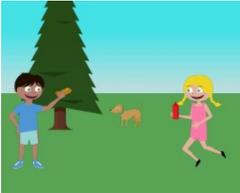

Ground truth
No

Human
-3.0058

Text baseline
-2.2792

Vision + Text
-2.5399

# Qualitative Results: Visual Paraphrasing

- Scenario 1: human, text baseline and our approach are all correct.

**Original Scene(s)**   **Descriptions**   **Generated Scenes**   **Answers**

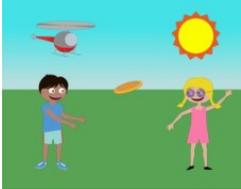

The bucket is in the sandbox. Mike runs to the ball. Mike is wearing a baseball cap.

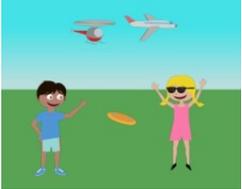

Ground truth
No

Human
-3.0058

Text baseline
-1.0911

Vision + Text
-1.3115

The bucket is in the sandbox. Mike runs to the ball. Mike is wearing a baseball cap.

# Qualitative Results: Visual Paraphrasing

- Scenario 2: human and our approach are correct while text baseline is incorrect

**Original Scene(s)**   **Descriptions**   **Generated Scenes**   **Answers**

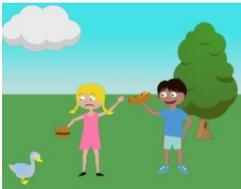

Mike is angry because Jenny won't play. Jenny is crying because Mike is mean. The owl watches the two children argue.

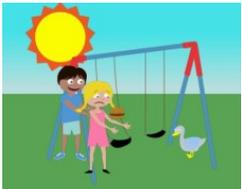

Ground truth
Yes

Human
1.3753

Text baseline
-0.1311

Vision + Text
0.2123

The helicopter is flying above Jenny. Mike wants Jenny's Frisbee. Jenny is crying because Mike is mad.

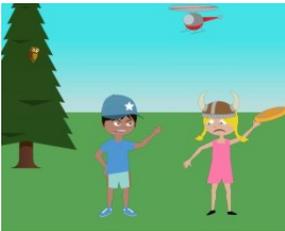

# Qualitative Results: Visual Paraphrasing

- Scenario 2: human and our approach are correct while text baseline is incorrect

| Original Scene(s) | Descriptions | Generated Scenes | Answers |
|---|---|---|---|

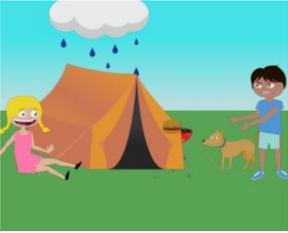

It is raining on the tent. Jenny is sitting on the ground. Mike is very mad.

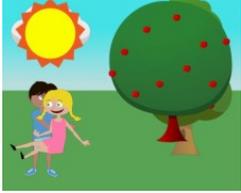

Jenny is sitting n the grass. Mike is angry with a dog. There is a burger on the grill

Ground truth
Yes

Human
2.7909

Text baseline
-0.1274

Vision + Text
0.2949

---

# Qualitative Results: Visual Paraphrasing

- Scenario 2: human and our approach are correct while text baseline is incorrect

| Original Scene(s) | Descriptions | Generated Scenes | Answers |
|---|---|---|---|

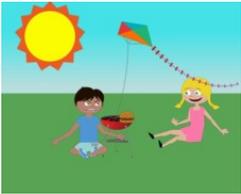

A lightening bolt flashes in the sky. Jenny is wearing a crown. Mike is shouting at Jenny.

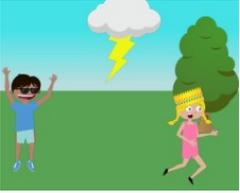

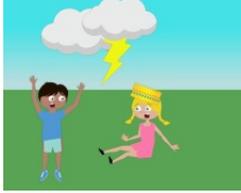

Jenny is singing on the swingset. Mike is happy to see Jenny at the park. The hot air ballon is high in the sky.

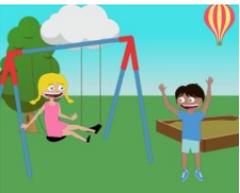

Ground truth
No

Human
-3.0058

Text baseline
0.2635

Vision + Text
-0.2044

# Qualitative Results: Visual Paraphrasing

- Scenario 2: human and our approach are correct while text baseline is incorrect

| Original Scene(s) | Descriptions | Generated Scenes | Answers |
|---|---|---|---|

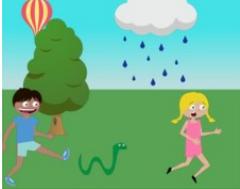

Jenny is running from a snake. Mike is chasing after the snake. It is raining on Jenny.

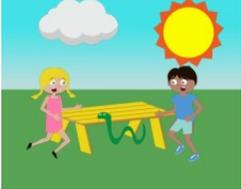

**Ground truth**
No

**Human**
-3.0058

**Text baseline**
0.1347

**Vision + Text**
-0.5795

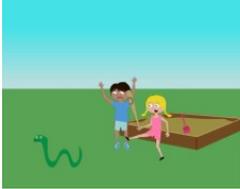

Jenny and Mike are afraid of the snake. Jenny is playing with a bat. Mike is jumping up.

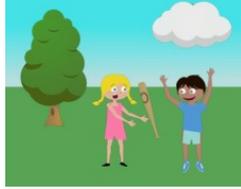

---

# Qualitative Results: Visual Paraphrasing

- Scenario 3: human and text baseline are correct while our approach is incorrect

| Original Scene(s) | Descriptions | Generated Scenes | Answers |
|---|---|---|---|

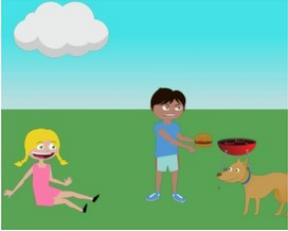

Mike and Jenny are having a barbecue. Jenny is excited to see a dog. Mike is angry at the dog for begging.

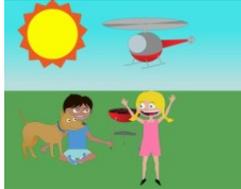

**Ground truth**
Yes

**Human**
1.3753

**Text baseline**
0.3909

**Vision + Text**
-0.1280

Jenny is sitting on the ground. Mike does not like his hamburger. The dog is wearing a blue collar

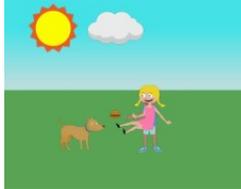

# Qualitative Results: Visual Paraphrasing

- Scenario 3: human and text baseline are correct while our approach is incorrect

| Original Scene(s) | Descriptions | Generated Scenes | Answers |
|---|---|---|---|

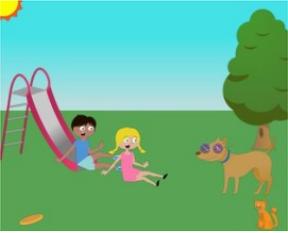

The cool dog is wearing sunglasses. The cat is jealous of the dog. Mike and Jenny play on the slide.

Mr. Dog is cool in sunglasses. Mike bumps into Jenny. Jenny is surprised by Mr. Dog.

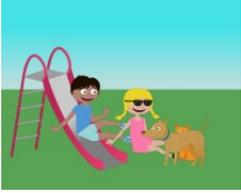

Ground truth
Yes

Human
1.3753

Text baseline
0.0509

Vision + Text
-0.6838

---

# Qualitative Results: Visual Paraphrasing

- Scenario 3: human and text baseline are correct while our approach is incorrect

| Original Scene(s) | Descriptions | Generated Scenes | Answers |
|---|---|---|---|

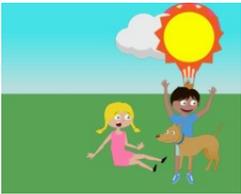

It is raining on Jenny. Mike wants Jenny's lunch. Jenny is giving Mike her wet lunch.

Jenny has a blue cap. Mike has a viking helmet. There are 2 trees.

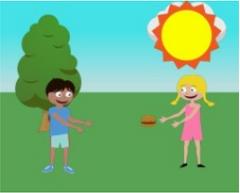

Ground truth
No

Human
-1.5452

Text baseline
-0.0278

Vision + Text
0.2061

15

# Qualitative Results: Visual Paraphrasing

- Scenario 3: human and text baseline are correct while our approach is incorrect

| Original Scene(s) | Descriptions | Generated Scenes | Answers |
|---|---|---|---|

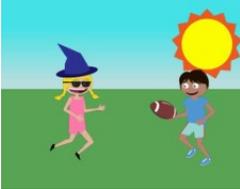

Jenny wears sunglasses Mike catches the football jenny is wearing a witch's hat

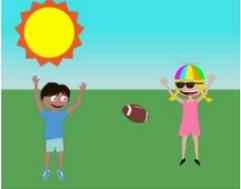

Mike is kicking the ball. Jenny wants to catch the ball. Jenny is smiling at Mike.

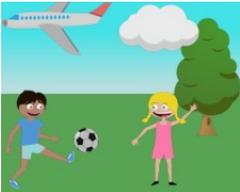

Ground truth
No

Human
-1.5452

Text baseline
-0.6850

Vision + Text
0.1486

# Qualitative Results: Visual Paraphrasing

- Scenario 4: human is correct while text baseline and our approach are incorrect

Original Scene(s)   Descriptions   Generated Scenes   Answers

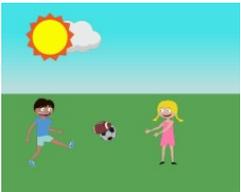

Mike is shooing the dag away. Jenny is waiting for a hamburger. The balloon flies over the playground.

Mike is cooking the burger. The dog is standing next to the pit. Jenny issitting in the grass.

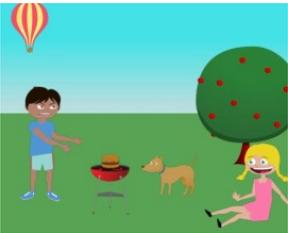

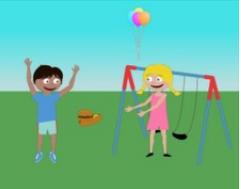

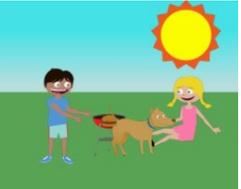

Ground truth
Yes

Human
4.2825

Text baseline
-0.1836

Vision + Text
-0.3634

# Qualitative Results: Visual Paraphrasing

- Scenario 4: human is correct while text baseline and our approach are incorrect

| Original Scene(s) | Descriptions | Generated Scenes | Answers |
|---|---|---|---|

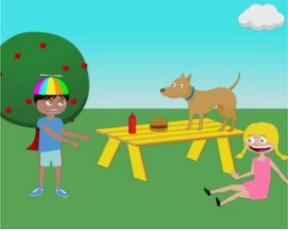

Mike is wearing a beanie cap. The dog wants to eat the hamburger. Jenny is happy to see Mike.

Mike is wearing a funny hat Jenny is laughing at Mike's hat Jenny is sitting next to the table

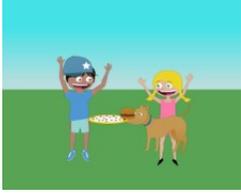

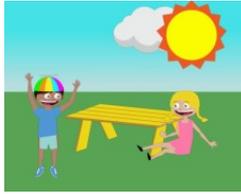

Ground truth
Yes

Human
2.7909

Text baseline
-0.4538

Vision + Text
-0.4682

---

# Qualitative Results: Visual Paraphrasing

- Scenario 4: human is correct while text baseline and our approach are incorrect

| Original Scene(s) | Descriptions | Generated Scenes | Answers |
|---|---|---|---|

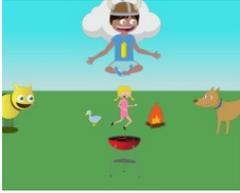

Jenny stood next to the fire. The dog watched the hamburgers on the grill. Mike flew into the sky with the mustard on his shirt.

Mike is near a grill. A dog is near jenny. there are three hot-dogs on the grill.

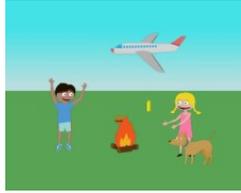

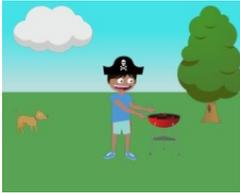

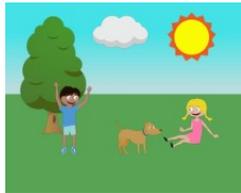

Ground truth
No

Human
-1.5452

Text baseline
1.7038

Vision + Text
1.2092

# Qualitative Results: Visual Paraphrasing

- Scenario 4: human is correct while text baseline and our approach are incorrect

| Original Scene(s) | Descriptions | Generated Scenes | Answers |
|---|---|---|---|

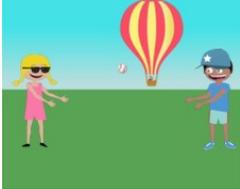

Mike is wearing a blue cap. Jenny is wearing a sunglasses. Jenny and Mike are playing catch.

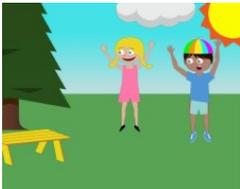

Ground truth
No

Human
-1.5452

Text baseline
0.5427

Vision + Text
0.2067

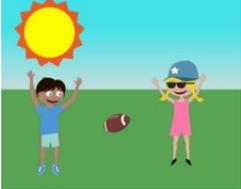

Mike is wearing a funny hat. Jenny is jumping off the ground. Mike is scared of something.

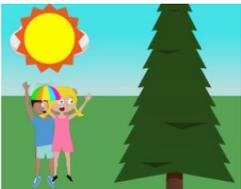

---

# Qualitative Results: Visual Paraphrasing

- Scenario 5: our approach and text baseline are correct while human is incorrect

| Original Scene(s) | Descriptions | Generated Scenes | Answers |
|---|---|---|---|

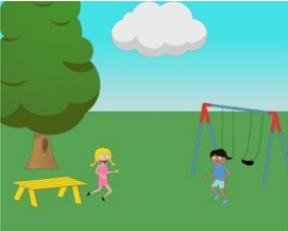

Mike is chasing Jenny. Jenny loves to play on the swings. The big tree is planted in the park.

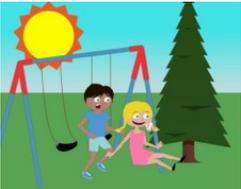

Ground truth
Yes

Human
-1.5452

Text baseline
0.5894

Vision + Text
0.6304

Jenny is running beside the table. Mike is running beside the swings. There is a cloud in the sky.

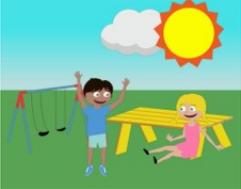

# Qualitative Results: Visual Paraphrasing

- Scenario 5: our approach and text baseline are correct while human is incorrect

| Original Scene(s) | Descriptions | Generated Scenes | Answers |

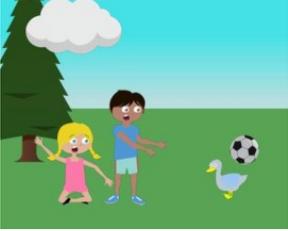

The duck is walking towards Mike and Jenny. Mike threw the soccer ball. Jenny is sitting in the grass.

Jenny and Mike are scared of the duck. Mr. Duck wants to help. Mike rolls the ball to Mr. Duck.

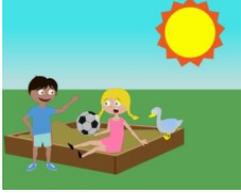

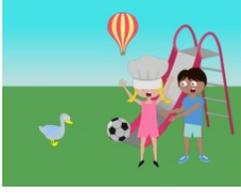

Ground truth
Yes

Human
-1.5452

Text baseline
0.8277

Vision + Text
1.2425

# Qualitative Results: Visual Paraphrasing

- Scenario 5: our approach and text baseline are correct while human is incorrect

| Original Scene(s) | Descriptions | Generated Scenes | Answers |

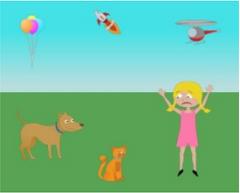

Jenny is upset. Jenny doesn't like cats. The dog will cheer Jenny up.

Jenny is crying by the cat and dog. Jenny is holding her hands out to the animals. There are balloons in the background.

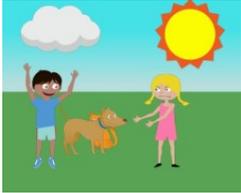

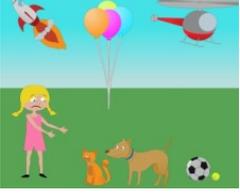

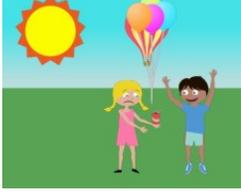

Ground truth
No

Human
1.3753

Text baseline
-0.0449

Vision + Text
-0.2418

# Qualitative Results: Visual Paraphrasing

- Scenario 5: our approach and text baseline are correct while human is incorrect

| Original Scene(s) | Descriptions | Generated Scenes | Answers |
|---|---|---|---|

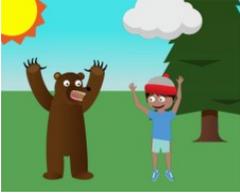

Mike is wearing a hat. The bear is roaring at Mike. Mike is in front of a tree.

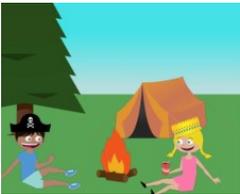

Ground truth
No

Human
1.3753

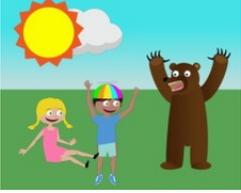

Mike is wearing a pirate hat. Jenny is wearing a crown. Jenny is holding her drink.

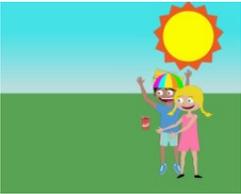

Text baseline
-1.1950

Vision + Text
-1.1451

---

# Qualitative Results: Visual Paraphrasing

- Scenario 6: our approach is correct while human and text baseline are incorrect

| Original Scene(s) | Descriptions | Generated Scenes | Answers |
|---|---|---|---|

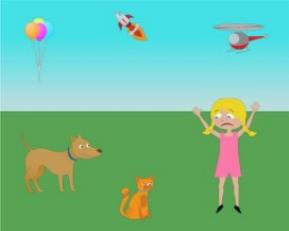

Jenny is upset. Jenny doesn't like cats. The dog will cheer Jenny up.

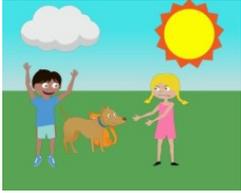

Ground truth
Yes

Human
-1.5452

The cat and dog are looking at Jenny. Jenny is looking at the animals and crying. There is a helicopter in the sky.

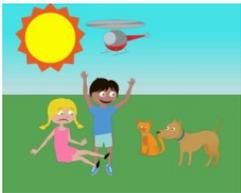

Text baseline
-0.0771

Vision + Text
0.6696

# Qualitative Results: Visual Paraphrasing

- Scenario 6: our approach is correct while human and text baseline are incorrect

| Original Scene(s) | Descriptions | Generated Scenes | Answers |
|---|---|---|---|

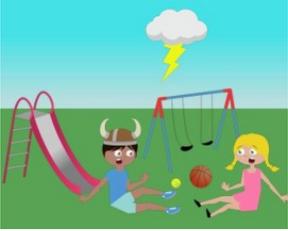

Mike and Jenny are sitting on the ground. Two balls are on the ground. Mike is next to the slide.

Jenny is sitting in the grass. Mike is wearing a vikings hat. Jenny is very surprised.

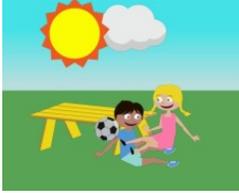

Ground truth
Yes

Human
-3.0058

Text baseline
-0.0863

Vision + Text
0.1524

# Qualitative Results: Visual Paraphrasing

- Scenario 6: our approach is correct while human and text baseline are incorrect

| Original Scene(s) | Descriptions | Generated Scenes | Answers |
|---|---|---|---|

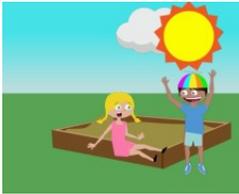

Mike is wearing a pirate hat. Jenny is wearing a funny hat. A dog is looking for something in the grass.

There is a rocket in the sky. Mike and Jenny are sitting on the ground. There is a dog in front of Mike and Jenny.

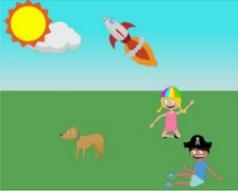

Ground truth
No

Human
1.3753

Text baseline
0.2037

Vision + Text
-0.0009

# Qualitative Results: Visual Paraphrasing

- Scenario 6: our approach is correct while human and text baseline are incorrect

| Original Scene(s) | Descriptions | Generated Scenes | Answers |
|---|---|---|---|

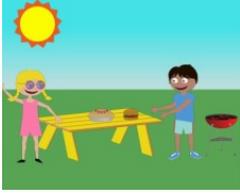

There's a pie on the table Jenny is wearing purple sunglasses Mike is beside the grill

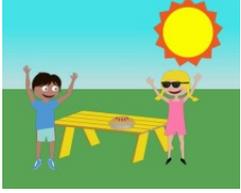

Ground truth
No

Human
1.3753

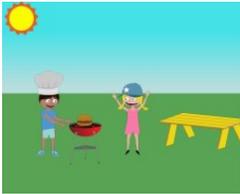

Mike put the hamburger onto the grill. Jenny was excited the hamburger was almost done. Mike cooked both hamburgers and hotdogs.

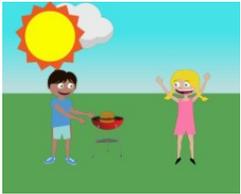

Text baseline
0.3193

Vision + Text
-0.1845

# Qualitative Results: Visual Paraphrasing

- Scenario 7: text baseline is correct while human and our approach are incorrect

| Original Scene(s) | Descriptions | Generated Scenes | Answers |
|---|---|---|---|

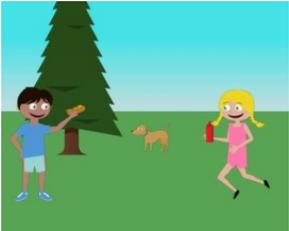

Mike is holding a hot dog Jenny is carring ketchup. Jenny is running.

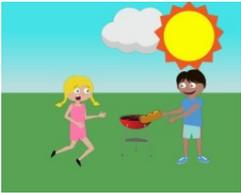

Ground truth
Yes

Human
-1.5452

Mike is very happy. Jenny is very happy. A dog is near a tree.

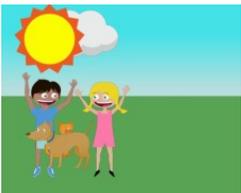

Text baseline
0.6291

Vision + Text
-0.1716

# Qualitative Results: Visual Paraphrasing

- Scenario 7: text baseline is correct while human and our approach are incorrect

| Original Scene(s) | Descriptions | Generated Scenes | Answers |
|---|---|---|---|

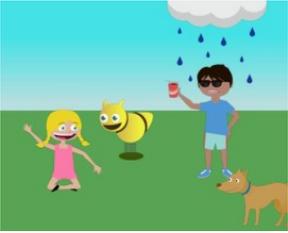

Rain is falling from the cloud. The dog is standing in front of Mike. Mike is wearing sunglasses.

Jenny is waving to Mike. Mike has a soda pop. It is raining today.

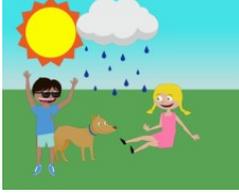

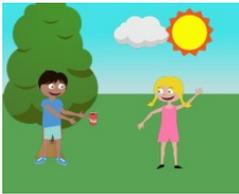

Ground truth
Yes

Human
-1.5452

Text baseline
0.0348

Vision + Text
-0.0688

---

# Qualitative Results: Visual Paraphrasing

- Scenario 7: text baseline is correct while human and our approach are incorrect

| Original Scene(s) | Descriptions | Generated Scenes | Answers |
|---|---|---|---|

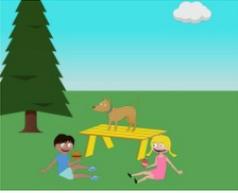

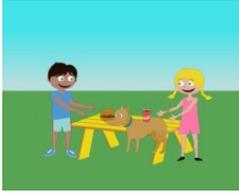

The dog is on the table. Mike has a hamburger. Jenny has a drink.

The plane is flying low. Mike likes hamburgers with ketchup. Jenny is laughing at Mike's joke.

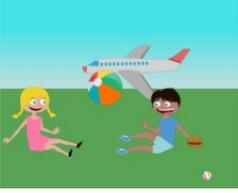

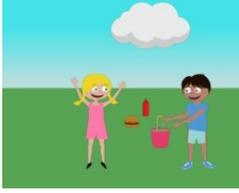

Ground truth
No

Human
1.3753

Text baseline
-0.3248

Vision + Text
0.1170

# Qualitative Results: Visual Paraphrasing

- Scenario 7: text baseline is correct while human and our approach are incorrect

| Original Scene(s) | Descriptions | Generated Scenes | Answers |
|---|---|---|---|

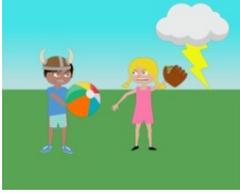

Lightning is coming out of the cloud. Mike and Jenny are angry. Mike is playing with a beach ball.

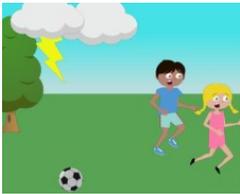

Ground truth
No

Human
1.3753

Text baseline
-0.0142

Vision + Text
0.8637

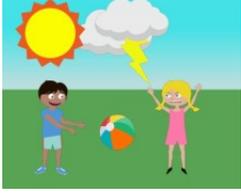

Mike and Jenny run away. Mike and Jenny are scared of lightening. Lightening is in the sky.

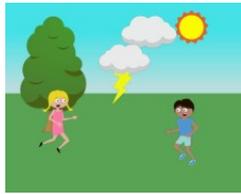

# Qualitative Results: Visual Paraphrasing

- Scenario 8: human, text baseline and our approach are all incorrect

| Original Scene(s) | Descriptions | Generated Scenes | Answers |
|---|---|---|---|

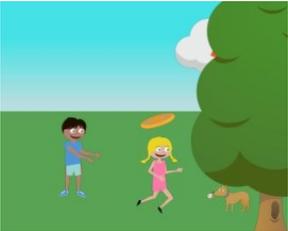

Mike is throwing the frisbee. Jenny is throwing the ball. The dog is standing next to the tree.

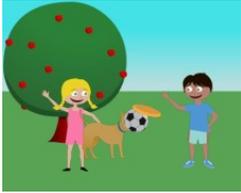

Ground truth
Yes

Human
-1.5452

Text baseline
-0.0217

Vision + Text
-0.3078

A dog has a baseball Jenny is running Mike is smiling

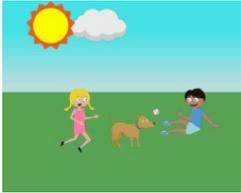

# Qualitative Results: Visual Paraphrasing

- Scenario 8: human, text baseline and our approach are all incorrect

Original Scene(s)   Descriptions   Generated Scenes   Answers

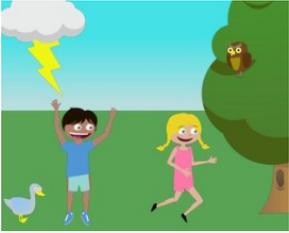

There is a lightning in the sky. Jenny is running from Mike. Mike is chasing Jenny.

A duck is near Mike An owl is in the tree. Lightning is coming out of the cloud.

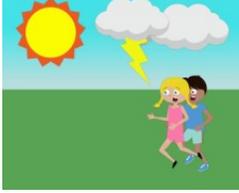

Ground truth
Yes

Human
-3.0058

Text baseline
-0.6132

Vision + Text
-0.3347

# Qualitative Results: Visual Paraphrasing

- Scenario 8: human, text baseline and our approach are all incorrect

Original Scene(s)   Descriptions   Generated Scenes   Answers

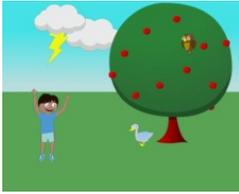

Mike and Jenny play on the swings. The dog watches Mike on the swing. The tall tree looks pretty.

Jenny is playing on the swing. The dog is standing next to mike. Mike is holding a burger.

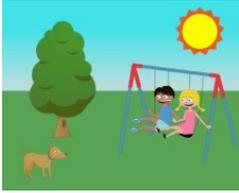

Ground truth
No

Human
1.3753

Text baseline
1.1652

Vision + Text
1.0543

25

# Qualitative Results: Visual Paraphrasing

- Scenario 8: human, text baseline and our approach are all incorrect

| Original Scene(s) | Descriptions | Generated Scenes | Answers |
|---|---|---|---|

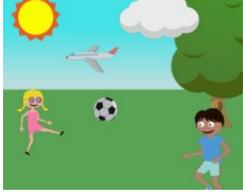

Jenny is kicking a ball. Jenny is wearing sunglasses. Mike is smiling.

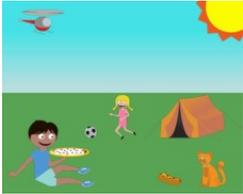

Ground truth
No

Human
4.2825

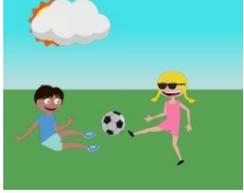

It is a sunny day. Mike is sitting with a pizza. Jenny is playing with a soccer ball.

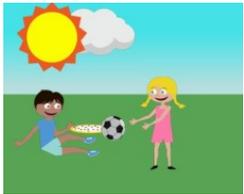

Text baseline
0.0234

Vision + Text
0.1555


\end{document}